\documentclass[pdflatex,sn-mathphys-num]{sn-jnl}

\usepackage{graphicx}%
\usepackage{multirow}%
\usepackage{amsmath,amssymb,amsfonts}%
\usepackage{amsthm}%
\usepackage{mathrsfs}%
\usepackage[title]{appendix}%
\usepackage{xcolor}%
\usepackage{textcomp}%
\usepackage{manyfoot}%
\usepackage{booktabs}%
\usepackage{subcaption}%
\usepackage{algorithm}%
\usepackage{algorithmicx}%
\usepackage{algpseudocode}%
\usepackage{listings}%
\usepackage{makecell}%
\newcommand{\angstrom}{\mbox{\normalfont\AA}}%
\usepackage{bm}%
\usepackage{lineno}%

\theoremstyle{thmstyleone}%
\theoremstyle{thmstyletwo}%

\theoremstyle{thmstylethree}%

\begin{document}


\title[Article Title]{A Generation Framework with Strict Constraints for Crystal Materials Design}


\author[1,2,3]{\fnm{Chao} \sur{Huang}}\email{chuang@ict.ac.cn}

\author[2]{\fnm{Jiahui} \sur{Chen}}
\equalcont{These authors contributed equally to this work.}

\author[2]{\fnm{Chen} \sur{Chen}}
\equalcont{These authors contributed equally to this work.}

\author[2]{\fnm{Chen} \sur{Chen}}

\author[2]{\fnm{Chunyan} \sur{Chen}}

\author[1,2]{\fnm{Panhao} \sur{Shen}}

\author[2,5]{\fnm{Renjie} \sur{Su}}

\author[4]{\fnm{Shiyu} \sur{Du}}

\affil[1]{\orgdiv{Institute of Computing Technology}, \orgname{Chinese Academy of Science}, \orgaddress{\city{Beijing}, \country{China}}}

\affil[2]{\orgname{Ningbo Institute of Artificial Intelligence Industry}, \orgaddress{\city{Ningbo}, \country{China}}}

\affil[3]{\orgdiv{Hangzhou Institute for Advanced Study}, \orgname{UCAS}, \orgaddress{\city{Hangzhou}, \country{China}}}

\affil[4]{\orgname{China University of Petroleum (East China)}, \orgaddress{\city{Qingdao}, \country{China}}}

\affil[5]{\orgdiv{Ningbo Institute of Materials Technology and Engineering}, \orgname{CAS}, \orgaddress{\city{Ningbo}, \country{China}}}



\abstract{The design of crystal materials plays a critical role in areas such as new energy development, biomedical engineering, and semiconductors. Recent advances in data-driven methods have enabled the generation of diverse crystal structures. However, most existing approaches still rely on random sampling without strict constraints, requiring multiple post-processing steps to identify stable candidates with the desired physical and chemical properties. In this work, we present a new constrained generation framework that takes multiple constraints as input and enables the generation of crystal structures with specific chemical  and properties. In this framework, intermediate constraints, such as symmetry information and composition ratio, are generated by a constraint generator based on large language models (LLMs), which considers the target properties. These constraints are then used by a subsequent crystal structure generator to ensure that the structure generation process is under control. Our method generates crystal structures with a probability of meeting the target properties that is more than twice that of existing approaches. Furthermore, nearly 100\% of the generated crystals strictly adhere to predefined chemical composition, eliminating the risks of supply chain during production.}




\maketitle
\newpage
\section{Introduction}\label{sec1}

Designing novel and high-performance crystal structures has long been recognized as a critical research direction in materials science. Traditional crystal materials design relies on computational simulations, such as Density Functional Theory (DFT)~\cite{orio2009density} and high-throughput calculations, which demand substantial human expertise and computational resources~\cite{yao2023machine}. Over the past few decades, the application of computer technology in the field of crystal structure generation has made systematic new materials design a reality~\cite{graser2018machine}, leading to the emergence of a series of representative methods, such as simulated annealing~\cite{timmermann2020iro}, random search~\cite{yang2021exploration}, genetic algorithms~\cite{cheon2020crystal}, and particle swarm optimization~\cite{2022CSPTHAC}. Nevertheless, these methods still exhibit limitations in generating valid and diverse crystal structures.

In recent years, large open-source material databases~\cite{jain2013commentary, schmidt2022large, schmidt2022dataset} have greatly advanced data-driven methods in crystal generation~\cite{nouira2018crystalgan,ren2022invertible}. Among these, generative models have been proved particularly effective in generating valid and diverse crystal structures~\cite{xie2021crystal,jiao2024crystal,miller2024flowmm,gruver2024fine,wu2025periodic}. However, constraint-based targeted crystal structure generation has remained challenging. CrystalFormer~\cite{cao2024space} generates space group-controlled crystal materials using space groups as initial constraint, while DiffCSP++~\cite{jiao2024space} produces strictly symmetry-compliant crystals based on elemental  and symmetry constraints. MatterGen~\cite{zeni2023mattergen} proposes a conditional diffusion model capable of guiding materials generation with specified constraints, including properties and chemical composition, though these constraints remain flexible and do not always fully meet the preset conditions.

In this work, we propose a novel two-step crystal structure generation framework with strict constraints, which we refer as CrystalGF. In the first step, our constraints generation module analyzes the input chemical composition and target material properties, then produces more strict constraints such as symmetry information and component ratios. In the second step, our structure generation module outputs the desired crystal structures conforming to both the original and newly generated constraints. Our contributions can be summarized as follows:

$\bullet$ We design a novel two-step crystal generation framework that only requires chemical composition and material properties as input information. By automatically generating symmetry-related constraints, this framework produces crystals that better satisfy target properties while strictly adhering to prescribed chemical composition.

$\bullet$ We develop a LLMs-based constraint generator that generates the space groups of unit cells, the Wyckoff positions for atoms, and the composition ratio based on specified chemical composition and target material properties. This module provides high-quality priors for the subsequent generation of structures and enables our methods to take input from natural language.

$\bullet$ We design a strict constrained crystal structure generator that operates under joint constraints of chemical composition, target properties, and symmetry requirements. This module ensures strict compliance with all specified constraints during structure generation.

$\bullet$ Compared to prior generative models, structures produced by our method are more than twice as likely to meeting the target properties. More precisely, 66.49\% of the generated structures achieved band gap deviations below 0.05 eV and 24.68\% exhibited formation energy deviations within 0.05 eV/atom. In addition, nearly 100\% of the generated crystals strictly adhere to the initial composition, eliminating supply chain risks during production.

\begin{figure*}[ht]
  \centering
  \includegraphics[width=\textwidth]{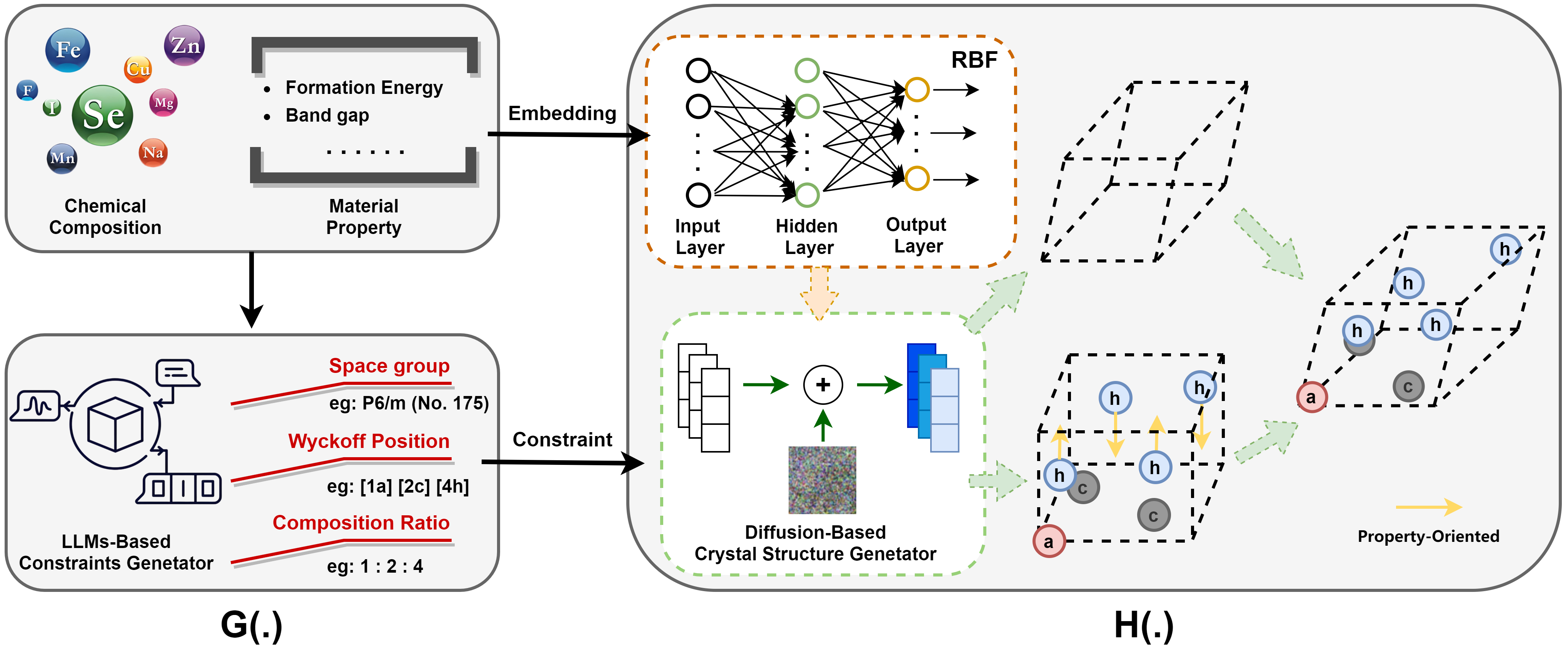}
  \caption{The schematic diagram of our proposed CrystalGF for the crystal structure generation. Where G(.) denotes the constraint generation module, responsible for parsing input data and generating additional constraints; H(.) represents the crystal structure generation module, which generates target crystal structures under multiple constraints.}
  \label{CrystalGF}
\end{figure*}

\section{Results}\label{sec2}

\subsection{Crystal Generation Framework}\label{subsec2.0}

In CrystalGF, we design a two-stage framework for generating crystal materials (Figure \ref{CrystalGF}). Many performance-oriented crystal generation methods overlook the influence of internal structural characteristics on material properties, whereas our approach establishes a connection between macroscopic performance and microstructure through a constraint generator (Appendix \ref{secB2}) and a structure generator (Appendix \ref{secB3}). Specifically, the constraint generator learns the relationship between the macroscopic properties and internal structural features of crystals to generate corresponding structural constraints, including space groups and Wyckoff positions of atoms. The structure generator, guided by multiscale constraints such as macroscopic properties and internal structural characteristics, produces crystal structures that meet the target performance criteria.

We conducted three sets of experiments to validate the effectiveness of our framework. First, to verify the impact of the constraint generator on final crystal structure generation, we selected four different generators and demonstrated a negative correlation between symmetry accuracy and property error (Section \ref{subsec2.1}). Second, we validated the effectiveness of our multi-conditional crystal structure generator (Section \ref{subsec2.2}). Finally, to confirm our framework's capability to accurately generate crystal materials satisfying target property requirements, we conducted comparative experiments with MatterGen and Con-CDVAE (Section \ref{subsec2.3}).

\begin{figure*}[ht]
  \centering
  \includegraphics[width=1\textwidth]{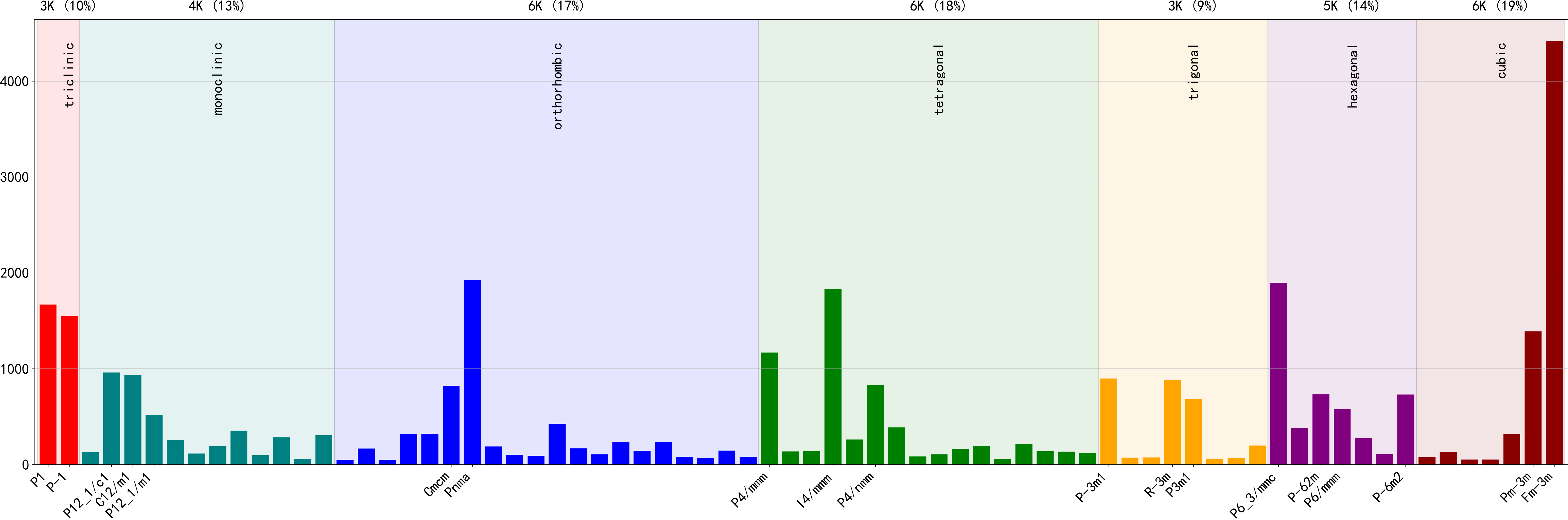}
  \caption{The distribution map of crystals across space group.}
  \label{spacegroup_bar}
\end{figure*}

\subsection{Crystal Constraints Generation}\label{subsec2.1}

We picked 34507 samples from the Materials Project (MP)~\cite{jain2013commentary} database, all of which were stable crystal structures, whose energy per atom after relaxation via DFT is within 0.1 eV per atom above the convex hull,  with less than 20 atoms in crystal unit cell. Each sample includes a crystal structure and its associated properties, such as band gap, formation energy, energy above hull.

\begin{table*}[ht]
    \centering
    
    \setlength{\tabcolsep}{1.6mm}{
    \begin{tabular}{cc|cccc}
        \toprule
        \multicolumn{2}{c|}{\textbf{Accuracy Rate}} & \multicolumn{2}{c}{\textbf{Space Group (\%)}} & \multicolumn{2}{c}{\textbf{Wyckoff Positions (\%)}} \\
        Input Pattern & Base LLM & Band. & Form. & Band. & Form. \\
        \midrule
        Element   & Qwen2.5-7B &  32.37 & 30.25 & 59.79 & 59.39 \\
        Composition & Llama-3.1-8B & \textbf{35.00} & \textbf{35.15} & \textbf{62.08} & \textbf{62.58} \\
        $\&$ Property & DeepSeek-R1-8B & 32.22 & 34.28 & 59.38 & 60.92 \\
        & GLM-4-9B & 33.61 & 32.48 & 60.24 & 60.14 \\
        \midrule
        Chemical & Qwen2.5-7B &  63.66 & 61.66 & 81.43 & 81.94 \\
        Formula & Llama-3.1-8B & \textbf{66.73} & \textbf{65.77} & 81.10 & 82.02 \\
        $\&$ Property & DeepSeek-R1-8B & 65.83 & 65.57 & 81.42 & 80.51 \\
        & GLM-4-9B & 64.76 & 64.36 & \textbf{82.83} & \textbf{82.13} \\
        \bottomrule 
    \end{tabular}}
    \caption{Accuracy rate on crystal symmetry generation.}
    \label{CrystalGF-L-results}
\end{table*}

\begin{figure*}[ht]
	\centering
	\begin{subfigure}{0.49\linewidth}
		\centering
		\includegraphics[width=\linewidth]
        {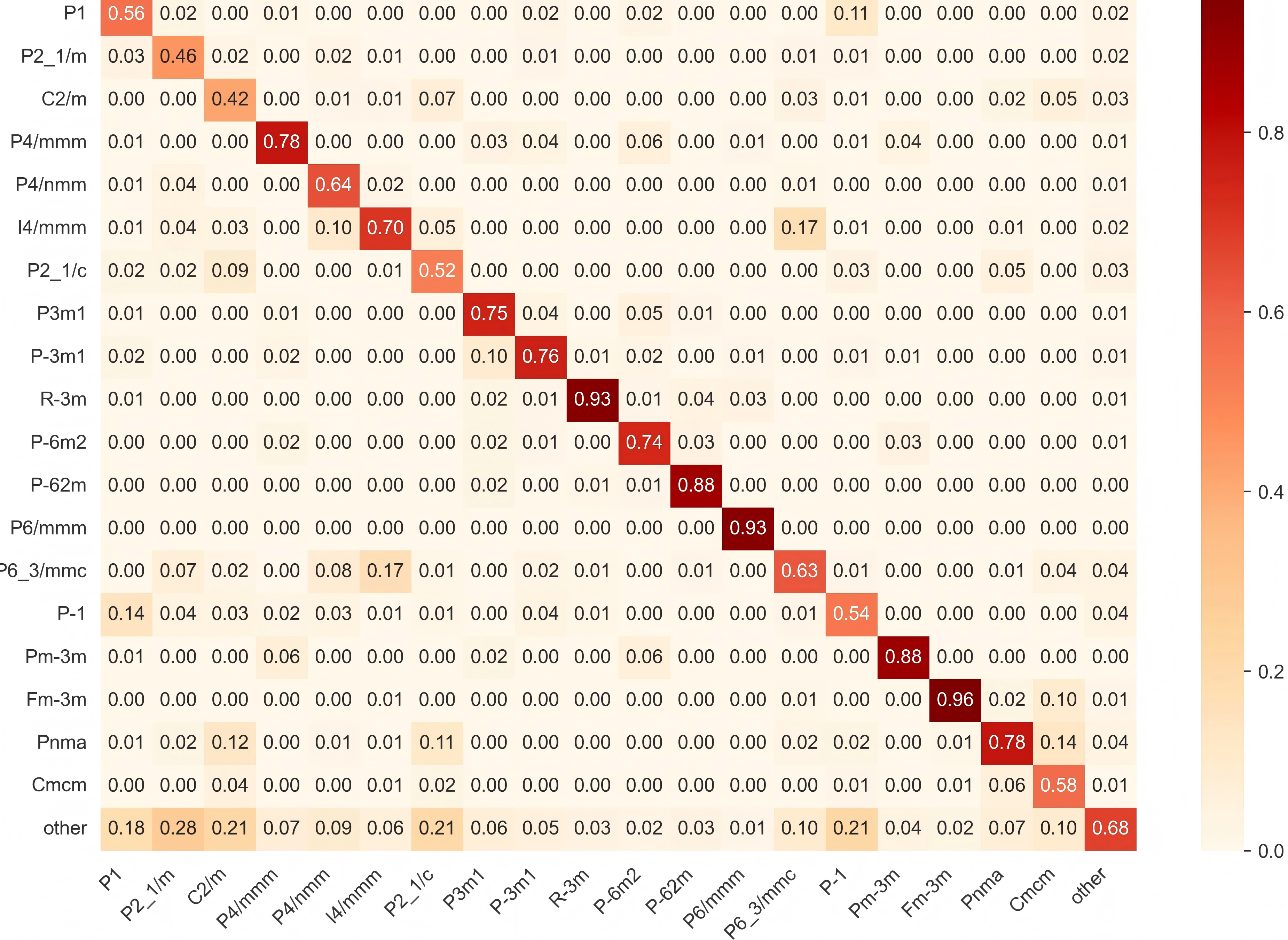}
		\caption{space groups}
		\label{space_group_bandgap}
	\end{subfigure}
	\centering
	\begin{subfigure}{0.49\linewidth}
		\centering
		\includegraphics[width=\linewidth]
        {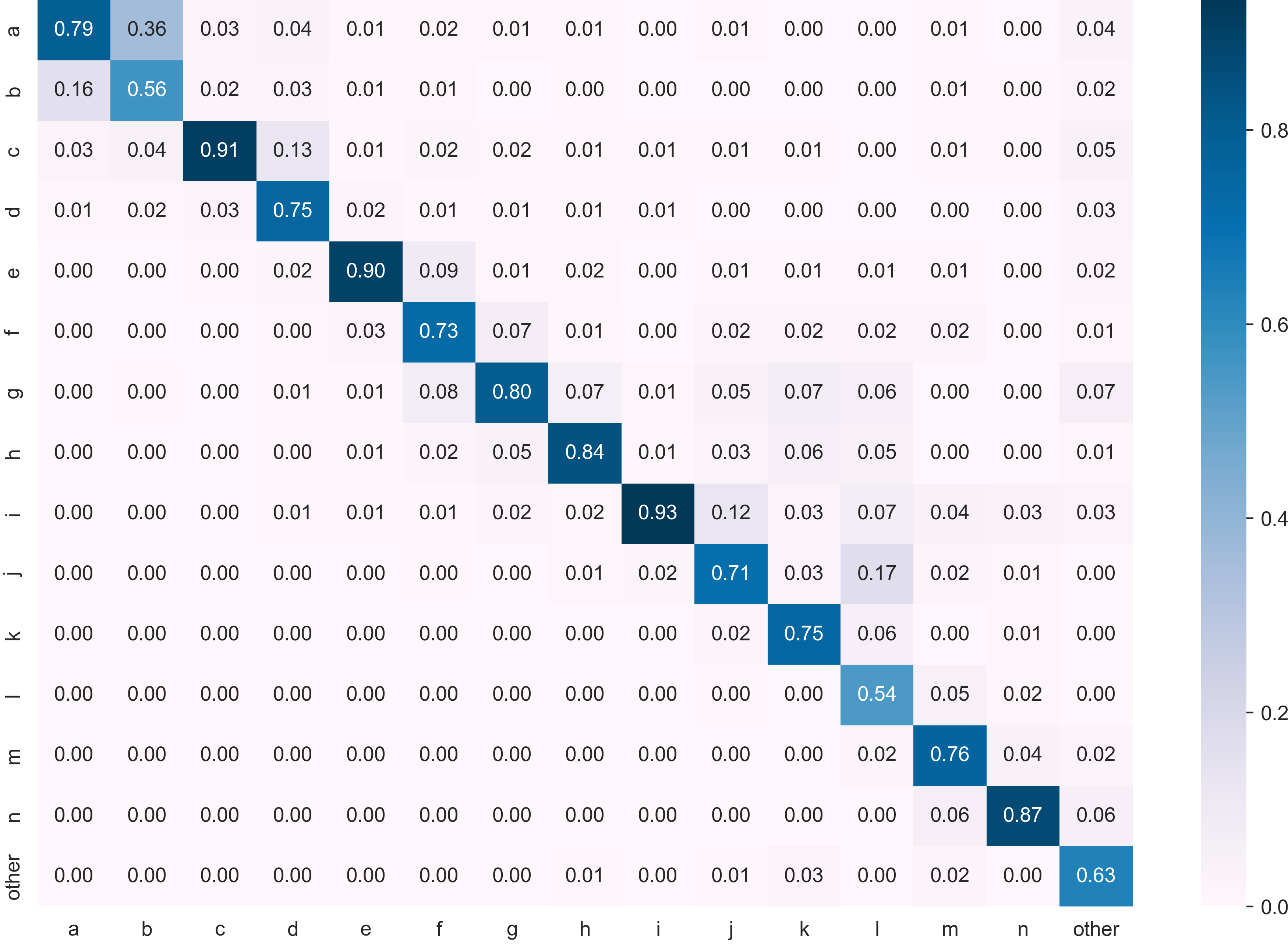}
		\caption{Wyckoff positions}
		\label{wyckoff_bandgap}
	\end{subfigure}
	\caption{Confusion matrix of space groups and Wyckoff positions generation results based on chemical formula and band gap. The horizontal and vertical axis indicate true labels and predicted labels, respectively.
    }
	\label{symmetry_predict}
\end{figure*}

The distribution of crystal space group types in our dataset is visualized in Figure \ref{spacegroup_bar}. Given the extreme sparsity for certain space group types, for instance, only 15 samples for $I2_12_12_1$ (No.24) and 0 sample for $Ia\text{-}3d$ (No.230), and the broad range of space group types, we adopted a strategy inspired by previous researchers~\cite{che2023crystal,wang2023crystallographic} to facilitate statistical analysis and presentation of model inference results: during model training, all original labels are preserved, but during evaluation, space group types with fewer than 500 instances are aggregated into an ``other'' category for statistical clarity. This approach is extended to Wyckoff positions generation, where Wyckoff letters with fewer than 50 occurrences are similarly categorized as ``other''.

We selected four widely downloaded open-source LLMs under 10B as the base models for the generator and fine-tuned them using the Llama-Factory~\cite{zheng2024llamafactory} framework. We provided more details in Appendix \ref{C1}. The results of the crystal symmetry generation are presented in Table \ref{CrystalGF-L-results}. In experiments using elemental composition and material property as input, the highest generation accuracies for space groups and Wyckoff positions reached 35.15\% and 62.58\% respectively, both obtained using the Llama-based generator. When using chemical formula and material property as input, the peak accuracies for space groups and Wyckoff positions improved to 66.73\% and 82.83\%, achieved by the Llama-based and GLM-based generators respectively. Correspondingly, when employing four constraint generators based on different LLMs and specifying bandgap and formation energy as property constraints, the error probability distributions between the simulated properties of crystals generated by CrystalGF and target properties are presented in Tables \ref{CrystalGF-element-error} and Table \ref{CrystalGF-formula-error}. Combined with the results from Table \ref{CrystalGF-L-results}, it demonstrates that higher symmetry accuracy achieved by constraint generators corresponds to smaller deviations between the generated crystal structures and target properties.

The confusion matrix of space groups and Wyckoff positions generation accuracy from the highest-precision constraint generator is illustrated in Figure \ref{symmetry_predict}. As exemplified in Figure \ref{space_group_bandgap}, there are five space group types that exceeded the 85\% generation accuracies: 88\% for $P\text{-}62m$ (No.189), 88\% for $Pm\text{-}3m$ (No.221), 93\% for $R\text{-}3m$ (No.166), 93\% for $P6/mmm$ (No.191), and 96\% for $Fm\text{-}3m$ (No.225). These space groups correspond to crystal systems with high symmetry. On the other hand, four space groups had generation accuracies below 55\%: 42\% for $C2/m$ (No.12), 46\% for $P2_1/m$ (No.11), 52\% for $P2_1/c$ (No.14), 54\% for $P\text{-}1$ (No.2). These space groups correspond to crystal systems with relatively low symmetry. We attributed this phenomenon to the fact that high-symmetry space groups (e.g., cubic crystal systems) possess greater numbers of symmetry operations and stringent geometric constraints, enabling models to more readily capture the correlations between their chemical composition and property parameters. In contrast, low-symmetry space groups (such as monoclinic or triclinic systems) exhibit complex structures with fewer symmetry elements and weaker parametric constraints. Minor variations in chemical composition or property metrics may lead to significant divergences in space group characteristics, thereby increasing the difficulty for models to learn these nonlinear relationships. In Figure \ref{wyckoff_bandgap}, the generation accuracy for Wyckoff positions of labels c, e, i, and n all exceeded 85\%. In contrast, the generation accuracy for labels b and l was relatively low but still remained around 55\%. This indicates that the model can efficiently capture the symmetric atomic structure within the unit cell when given the chemical composition, target properties, and space group information.

\subsection{Crystal Structure Generation}\label{subsec2.2}

We evaluated our crystal structure generation module on three publicly available datasets with distinct data distributions: Perov-5, MP-20, and MPTS-52. Perov-5 ~\cite{castelli2012new} comprises 18,928 perovskite crystals exhibiting structural similarity but compositional diversity, each containing 5 atoms per unit cell. MP-20 ~\cite{jain2013materials} contains 45,231 samples representing experimentally realized crystals with $\leq$ 20 atoms per unit cell. MPTS-52 serves as a more challenging extension of MP-20, consisting of 40,476 structures with up to 52 atoms per unit cell. For all datasets, we adopt the data splitting protocols established in prior works ~\cite{xie2021crystal,jiao2024crystal} to ensure benchmarking consistency. The hyper-parameters and training details are summarized in Appendix \ref{C1}.

For each generated crystal structure, we calculate the match rate through the Structure Matcher class in Pymatgen~\cite{ong2013python}, with thresholds stol = 0.5, angle tol = 10, ltol = 0.3, following previous work. The match rate represents the ratio of matched structures relative to the total number within the testing set, and the RMSE is averaged over the matched pairs, and normalized by $ \sqrt[3]{V/N}$, where $V$ is the volume of the lattice, $N$ is the atom number of crystals. We compared our approach with the following two generation-based methods, including CDVAE~\cite{xie2021crystal} and DiffCSP++~\cite{jiao2024space}. The results of CDVAE and DiffCSP++ are obtained by reproducing the inference process using publicly released datasets and pretrained model weights. 

As shown in Table \ref{CrystalGF-D-BF}, CrystalGF-DB and CrystalGF-DF denote models incorporating band gap and formation energy as input properties, respectively. Due to the absence of formation energy in Perov-5 dataset and band gap in MPTS-52 dataset, missing values are represented by dashes ``-''. For Perov-5 dataset, our method achieves comparable performance to DiffCSP++ with matching rate of 96.27\% (vs. 96.49\%) and RMSE of 0.0333 (vs. 0.0343). Regarding MP-20 dataset, our approach demonstrates marginal improvements where CrystalGF-DF reaches 84.00\% (vs. 83.18\%) matching rate and 0.0365 (vs. 0.0351) RMSE. Notably on MPTS-52 dataset, we observe more significant enhancements with 45.38\% (vs. 44.20\%) matching rate and 0.0431 (vs. 0.0451) RMSE. Experimental results validate that our module achieves more pronounced improvements on complex tasks, which can be attributed to both the information fusion capability of the multi-head cross-attention mechanism and the property-oriented generation capacity of module in low symmetry environments enabled by property embedding optimization. In addition, the band gap based model exhibits certain performance degradation in Perov-5 dataset, which we speculated stems from the excessive sparsity of input feature representations caused by numerous band gap values approaching zero, leading to suboptimal results.

\begin{table*}[ht]
    \centering
    
    \setlength{\tabcolsep}{2.8mm}{
    \begin{tabular}{ccccccc}
        \toprule
        {} & \multicolumn{2}{c}{Perov-5} & \multicolumn{2}{c}{MP-20} & \multicolumn{2}{c}{MPTS-52}\\
        {} & MR(\%) & RMSE & MR(\%) & RMSE & MR(\%) & RMSE \\
        \midrule
        CDVAE & 48.58 & 0.1009 & 35.66 & 0.1020 & 6.16 & 0.1904 \\
        DiffCSP++ & \textbf{96.49} & 0.0343 & 83.18 & 0.0351 & 44.20 & 0.0451\\
        \midrule
        CrystalGF-DB & 96.27 & \textbf{0.0333} & 83.20 & \textbf{0.0345} & - & -\\
        CrystalGF-DF & - & - & \textbf{84.00} & 0.0365 & \textbf{45.38} & \textbf{0.0431} \\
        \bottomrule 
    \end{tabular}}
    \caption{Results on crystal structure generation. MR stands for Match Rate.}
    \label{CrystalGF-D-BF}
\end{table*}

\subsection{Structures Metrics Calculation}\label{subsec2.3}

We employ the structural metrics established by CDVAE~\cite{xie2021crystal} and the S.U.N. evaluation tool provided by MatterGen~\cite{zeni2023mattergen} to compare different methods. The evaluation metrics from CDVAE~\cite{xie2021crystal} include empirical validity, coverage of the test set, and the Wasserstein distances between the distributions of three properties of the generated samples and the test set: atomic density ($ d_\rho $), number of unique elements ($ d_{elem} $), and predicted formation energy ($ d_E $). For each method, we generate 10,000 samples for evaluation. The results are presented in Table \ref{cdvae-metrics}. CrystalGF performs comparably to other methods across most metrics, while exhibiting a higher probability of generating valid crystal structures, which can be attributed to its high-quality constrained priors and precise structure generation.

\begin{table*}[ht]
    \centering
    
    \setlength{\tabcolsep}{2.2mm}{
    \begin{tabular}{lccccccc}
        \toprule
         & \multicolumn{2}{c}{Validity (\%) (↑)} & \multicolumn{2}{c}{Coverage (\%) (↑)} & \multicolumn{3}{c}{Property Distribution (↓)}\\
            MP-20 & Struct. & Comp. & Recall & Precision & $d_\rho$ & $d_E$ & $d_{elem}$  \\
        \midrule
        CDVAE & 99.97 & 85.61 & 99.31 & 99.47 & 0.70 & 0.24 & 1.28  \\
        DiffCSP++ & 99.44 & 86.50 & 99.72 & 99.61 & 0.12 & 0.05 & 0.33  \\
        FlowMM & 96.67 & 83.25 & 99.49 & 99.58 & 0.23 & 0.09 & 0.08  \\
        SymmCD & 90.34 & 85.81 & 99.58 & 97.76 & 0.23 & 0.21 & 0.40  \\
        Con-CDVAE & 99.97  & 81.77 & 98.29 & 99.39 & 0.67 & 0.15 & 0.73 \\
        MatterGen & 99.56 & 84.50 & 99.50 & 99.33 & 0.26 & 0.17 & 0.11 \\
        CrystalGF & 99.85 & 90.76 & 99.31 & 99.65 & 0.17 & 0.07 & 0.09\\
        \bottomrule 
    \end{tabular}}
    \caption{Results for comparing the validity, coverage, and property distribution metrics.}
    \label{cdvae-metrics}
\end{table*}

Stable, unique and novel (S.U.N.) is a critical metric for evaluating the efficiency and cost-effectiveness in discovering new materials during model inference. In this study, to reduce the computational cost of structural relaxation, we use the validation tools provided by MatterGen to compute the S.U.N. values. Thermodynamic stability is determined by calculating the convex hull energy of the materials, where the convex hull represents the lowest-energy mixing state of the material. We employ a pre-trained mattersim\cite{yang2024mattersim} model to predict the energy of generated crystals and run structural relaxation to achieve higher stability. Novelty validation is conducted using the Alex-MP-ICSD reference dataset. According to our research, the crystal system has a significant impact on the S.U.N. metrics. Highly symmetric crystal systems, such as hexagonal and cubic, possess abundant symmetry operations, exhibit greater structural stability, and occur more frequently in nature. In contrast, crystal systems with lower symmetry, such as triclinic and monoclinic, have fewer symmetry operations and looser structural constraints, making them more prone to generating novel structures. However, crystals with low-symmetry space groups are relatively scarce in nature and pose greater synthetic challenges. even if such crystals could be designed theoretically, their practical utility remains limited. Therefore, to balance the influence of different crystal systems, we proposed a balanced S.U.N. metrics, the calculation formula is presented below:

\begin{equation}\label{wsun}
    \text{S.U.N.}_{\text{balanced}} = \frac{1}{k} \sum_{c \in S} \text{S.U.N.}_c
\end{equation}

where $k$ denotes the number of crystal systems with generated structures, and $S$ represents the set of all crystal systems. For each crystal system $c \in S$, $\text{S.U.N.}_{c}$ denotes the S.U.N. score of all structures within that crystal system.

\begin{figure*}[ht]
  \centering
  \includegraphics[width=1\textwidth]{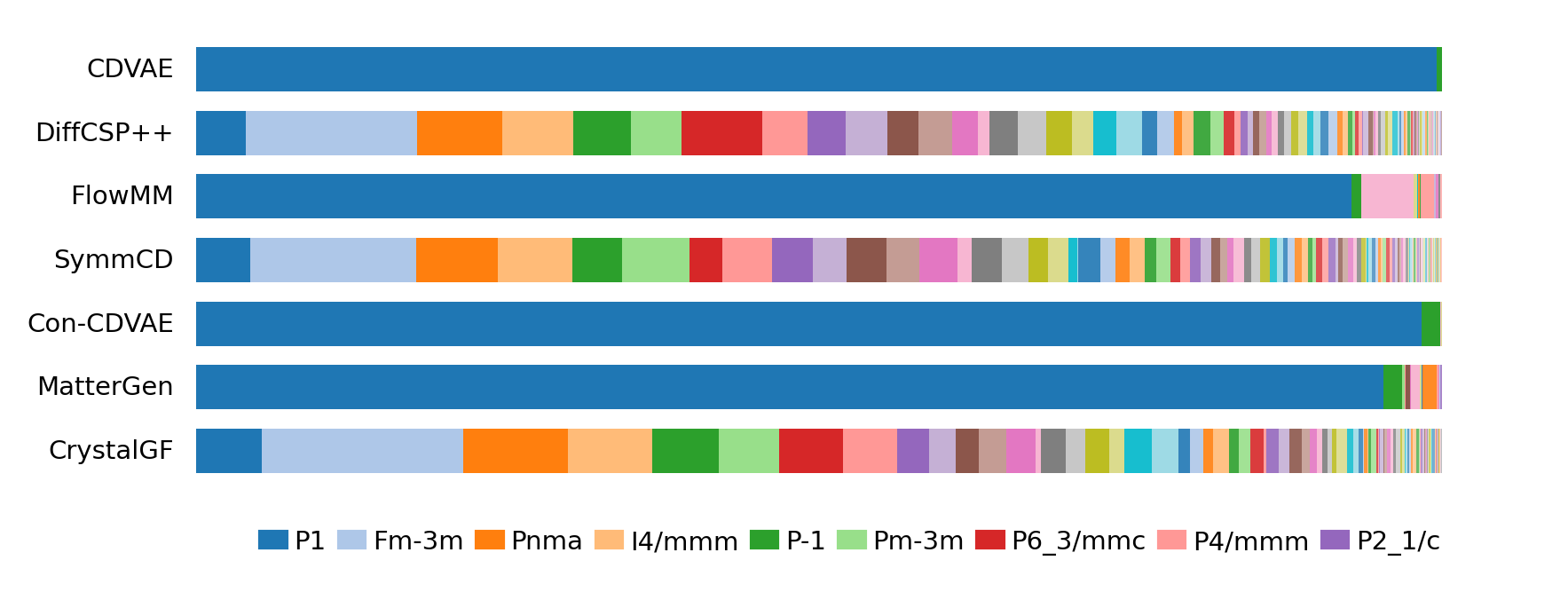}
  \caption{Proportions of space groups for structures generated by each method. The width of each colored segment represents the proportion of crystals possessing that symmetry.}
  \label{method_spg_dist}
\end{figure*}

As shown in Table \ref{S.U.N.}, CrystalGF outperforms SymmCD and FlowMM in terms of the S.U.N. metric. Due to the strong constraint framework adopted in our approach, the generated crystals adheres to the structural characteristics (e.g., symmetry) of many existing crystals. As a result, compared with methods such as DiffCSP++ and Symmcd, CrystalGF produces structures with relatively lower novelty, while exhibiting a significant advantage in stability. Furthermore, both Con-CDVAE and MatterGen demonstrate substantially superior S.U.N. performance compared to other methods. However, in the balanced S.U.N. metric, our CrystalGF exhibit minor improvements, whereas Con-CDVAE and MatterGen demonstrate more pronounced decreases. Combined with the space group distribution of crystal structures generated by each method, as illustrated in Figure \ref{method_spg_dist}, it is evident that a significant proportion of the structures generated by Con-CDVAE and MatterGen belong to space group 1. Although their overall S.U.N. metrics are relatively high, the lack of symmetry constraints limits their ability to discover S.U.N. high-symmetry structures. In contrast, CrystalGF exhibit a more uniform distribution across space groups from 1 to 230, thereby enabling the exploration of novel structures across a broader range of structural characteristics.

\begin{table*}[ht]
    \centering
    
    \setlength{\tabcolsep}{2.5 mm}{
    \begin{tabular}{lccccc}
        \toprule
         \multirow{2}{*}{Method} & \multirow{2}{*}{Unique} &  \multirow{2}{*}{Novel} & Relaxed & Relaxed & Balanced \\
         {} & {} & {} & Stable & S.U.N. & S.U.N. \\
        \midrule
        CDVAE & 99.37\% & 71.66\% & 38.43\% & 14.45\% & 7.23\% \\
        DiffCSP++ & 98.57\% & 32.82\% & 70.90\% & 9.55\% & 11.17\% \\
        FlowMM & 83.35\% & 93.43\% & 9.64\% & 4.18\% &  1.74\% \\
        SymmCD & 97.73\% & 74.23\% & 30.34\% & 8.37\% & 8.63\% \\
        Con-CDVAE & 91.15\% & 76.81\% & 46.09\% & 25.03\% & 12.53\% \\
        MatterGen & 98.97\% & 78.21\% & 46.19\% & 27.02\% & 7.58\% \\
        CrystalGF & 96.87\% & 39.67\% & 64.67\% & 9.38\% & 11.14\% \\
        \bottomrule 
    \end{tabular}}
    \caption{Percent of S.U.N. samples produced from 10,000 generated crystals for each method.}
    \label{S.U.N.}
\end{table*}

\subsection{Material Properties Comparison}\label{subsec2.4}

In this subsection, we compared the differences between the simulated properties of the generated structures and the target properties. We employ density functional theory (DFT) for computational simulations using the Vienna Ab initio Simulation Package~\cite{kresse1996efficient}~\cite{kresse1996efficiency} (VASP), more details in Appendix \ref{C3}. For comparative analysis, we fine-tuned and retrained two multi-conditional crystal generation models: MatterGen and Con-CDVAE, the training details are provided in the Appendix \ref{C1}. In practical crystal materials design, it is considered acceptable if the absolute error of the properties is less than 0.05, and meet the requirement if less than 0.01, and the unit of band gap and formation energy are eV and eV/atom, respectively. Therefore, we compared the probability distributions of property absolute errors between different models in Tables \ref{CrystalGF-element-error} and Table \ref{CrystalGF-formula-error}.

Our framework benefits from the precise symmetry and composition ratio generation of the constraint generator, along with the strict multi-conditional constraints imposed by the structure generator, enabling higher precision of generating crystal structures with target inputs. 

In generation tasks targeting elemental composition and material properties, our method demonstrates improvements over MatterGen in both property and composition constraint capabilities. Specifically, CrystalGF, when using LLaMA as the base constraint-aware generator, achieves a 10\% improvement in elemental precision compared to MatterGen. Moreover, structures generated by our approach are more than twice as likely to satisfy the target properties relative to prior generative models. In particular, CrystalGF doubles the success rate of MatterGen in generating structures with formation energy differences below 0.01 eV/atom.

For generation tasks targeting chemical formulas and material properties, Specifically, it achieves nearly 100\% compliance with elemental constraints, whereas Con-CDVAE attains an average probability of only 5.81\%. This disparity is primarily attributed to Con-CDVAE’s limitation of modeling only structural and property distributions during its initial training phase. In terms of property constraints, CrystalGF, when using GLM as the constraint-aware generative base model, achieves approximately 30\% higher probability than Con-CDVAE in generating structures with formation energy differences below 0.05 eV/atom.

We present partially generation results exhibiting high similarity to structures in the test set in Figures \ref{same}. Owing to precise symmetry constraint generation and the superior accuracy of our structure generator trained on the MP dataset, the generated structures closely resemble their MP counterparts, resulting in nearly identical computational material properties. Meanwhile, Figure \ref{diff} showcases generated structures that deviate more noticeably from the test set. Although these structures exhibit partial discrepancies in symmetry constraint generations compared to original data, their computed properties remain proximate to target values. Notably, no analogous structures exist in the MP dataset, highlighting the capability of our framework not only to accurately reproduce known materials but also to discover novel ones.

\begin{table*}[ht]
    \centering
    
    \setlength{\tabcolsep}{2.2mm}{
    \begin{tabular}{lc|ccc|ccc}
        \toprule
        {} & {} & \multicolumn{2}{c}{Prob. of BG (\%)} & \multirow{2}{*}{Comp.(\%)} &\multicolumn{2}{c}{Prob. of FE (\%)} & \multirow{2}{*}{Comp.(\%)}  \\ Method & Error Bars & \textless{} 0.01 & \textless{} 0.05 & {} & \textless{} 0.01 & \textless{} 0.05 \\ 
        \midrule
        \multicolumn{2}{l|}{MatterGen} & 48.69 & 63.03 & 89.80 & 3.30 & 15.90 & 84.28 \\
        \multicolumn{2}{l|}{CrystalGF (Qwen)} & 48.29 & \textbf{66.49} & \textbf{100.00} & 5.73 & 20.85 & 99.96 \\    
        \multicolumn{2}{l|}{CrystalGF (DS)} & 48.96 & 65.92 & 99.98 & 6.41 & 23.67 & 99.96 \\
        \multicolumn{2}{l|}{CrystalGF (GLM)} & 48.07 & 66.27 & 99.98 & 6.37 & 22.86 & 99.97 \\
        \multicolumn{2}{l|}{CrystalGF (Llama)} & \textbf{48.87} & 66.38 & 99.97 & \textbf{6.81} & \textbf{24.68} & \textbf{99.98} \\
        \bottomrule
    \end{tabular}}
    \caption{The probability distribution of property absolute errors and precision of chemical composition of crystal structures generated by elemental composition and material properties.}
    \label{CrystalGF-element-error}
\end{table*}

\begin{table*}[ht]
    \centering
    
    \setlength{\tabcolsep}{2.2mm}{
    \begin{tabular}{lc|ccc|ccc}
        \toprule
        {} & {} & \multicolumn{2}{c}{Prob. of BG (\%)} & \multirow{2}{*}{Comp.(\%)}&\multicolumn{2}{c}{Prob. of FE (\%)} & \multirow{2}{*}{Comp.(\%)}  \\ Method & Error Bars & \textless{} 0.01 & \textless{} 0.05 & {} & \textless{} 0.01 & \textless{} 0.05 \\ 
        \midrule
        \multicolumn{2}{l|}{Con-CDVAE} & \textbf{56.13} & 64.33 & 7.02 & 0.63 & 2.92 & 4.33 \\
        \multicolumn{2}{l|}{CrystalGF (Qwen)} & 49.78 & 67.61 & \textbf{100.00} & 11.28 & 32.68 & 99.99 \\    
        \multicolumn{2}{l|}{CrystalGF (Llama)} & 49.64 & 67.64 & \textbf{100.00} & 11.71 & \textbf{33.90} & 99.98 \\
        \multicolumn{2}{l|}{CrystalGF (DS)} & 49.82 & 67.88 & \textbf{100.00} & 11.43 & 32.59 & 99.99 \\
    \multicolumn{2}{l|}{CrystalGF (GLM)} & 49.47 & \textbf{67.96} & \textbf{100.00} & \textbf{12.02} & 33.74 & \textbf{100.00} \\
        \bottomrule
    \end{tabular}}
    \caption{The probability distribution of property absolute errors and precision of chemical composition of crystal structures generated by chemical formula and material properties.}
    \label{CrystalGF-formula-error}
\end{table*}

\begin{appendices}

\section{Crystal Supplementary Knowledge}\label{secA}

\subsection{Crystal Structure representation}\label{secA1}

Any crystal structure $M$ can be described as the periodic arrangement of atoms in 3D space. The repeating unit is called a unit cell. A unit cell that includes $n$ atoms can be fully described by a triplet, denoted as $\left(A, X, L\right)$, where $ A = \left[ a_1,a_2,...,a_n \right] \in \mathbb{R}^{h \times n}$ represents the one-hot representations of atom type, $ X = \left[ x_1,x_2,...,x_n \right] \in \mathbb{R}^{3 \times n}$ comprises the coordinates of atoms, and $ L = \left[ l_1,l_2,l_3 \right] \in \mathbb{R}^{3 \times 3}$ is the lattice matrix containing three basic vectors to periodically translate the unit cell to the entire 3D space. The infinite periodic structure $ M = \left(A, X, L\right)$ can be represented as

\begin{equation} 
    M := \left\{ \left( a_i,x_i^{'} \right) | x_i^{'} = x_i + \sum_{i=1}^3 k_il_i, k_i \in \mathbb{Z} \right\}
\label{crystal structure}
\end{equation}

\noindent where $k_i$ are any integers that translate the unit cell using $L$ to tile the entire 3D space.

\subsection{Crystal Symmetry Constraint}\label{secA2}

Crystal symmetry, encompassing space groups and Wyckoff positions, plays a pivotal role in the description of crystal structures. We illustrates the influence of space group types on lattice parameters in Table \ref{spg2lattice}. Space group types are numbered from 1 to 230 and correspond to the seven crystal systems. Among them, Space groups with higher numerical indices impose more stringent lattice constraints. For instance, 195 $\sim$ 230 belong to the cubic system, where the lattice adopts a cube-shaped configuration with only one degree of freedom in lattice parameters. Conversely, 1$\sim$ 2 correspond to the triclinic system, featuring lattices of arbitrary geometry with six degrees of freedom. The trigonal system presents a unique case, incorporating both rhombohedral and hexagonal lattice configurations, thereby encompassing two distinct types of lattice constraints.

\begin{table}[ht]
  \caption{The constraint of space group types on lattice shape, where a, b, c represent the lengths of the lattice vectors, and $\alpha$, $\beta$, $\gamma$ represent the angles between them.}
  \label{spg2lattice}
  \centering
  \begin{tabular}{cccc}
    \toprule
    Crystal System & Space Group & Length Constraint & Angle Constraint\\
    \midrule
    Triclinic & 1 $ \sim $ 2 & $a \neq b \neq c$ & $\alpha \neq \beta \neq \gamma$\\
    \midrule
    Monoclinic & 3 $ \sim $ 15 & $a \neq b \neq c$ & $\alpha = \gamma = 90^\circ , \beta \neq 90^\circ$ \\
    \midrule
    Orthorhombic & 16 $ \sim $ 74 & $a \neq b \neq c$ & $\alpha = \beta = \gamma = 90^\circ$ \\
    \midrule
    Tetragonal & 75 $ \sim $ 142 & $ a = b \neq c $ & $\alpha = \beta = \gamma = 90^\circ$ \\
    \midrule
    Trigonal & 143 $ \sim $ 167 & \makecell{$ a = b = c $ \\ $ a = b \neq c $} & \makecell{$ \alpha = \beta = \gamma \neq 90^\circ $ \\ $ \alpha = \beta = 90^\circ , \gamma=120^\circ $}\\
    \midrule
    Hexagonal & 168 $ \sim $ 194 & $ a = b \neq c $ & $ \alpha = \beta = 90^\circ , \gamma=120^\circ $ \\
    \midrule
    Cubic & 195 $ \sim $ 230 & $ a = b = c $ & $\alpha = \beta = \gamma = 90^\circ$ \\
    \bottomrule 
  \end{tabular}
\end{table}

\begin{figure}[ht]
  \centering
  \includegraphics[width=1\textwidth]{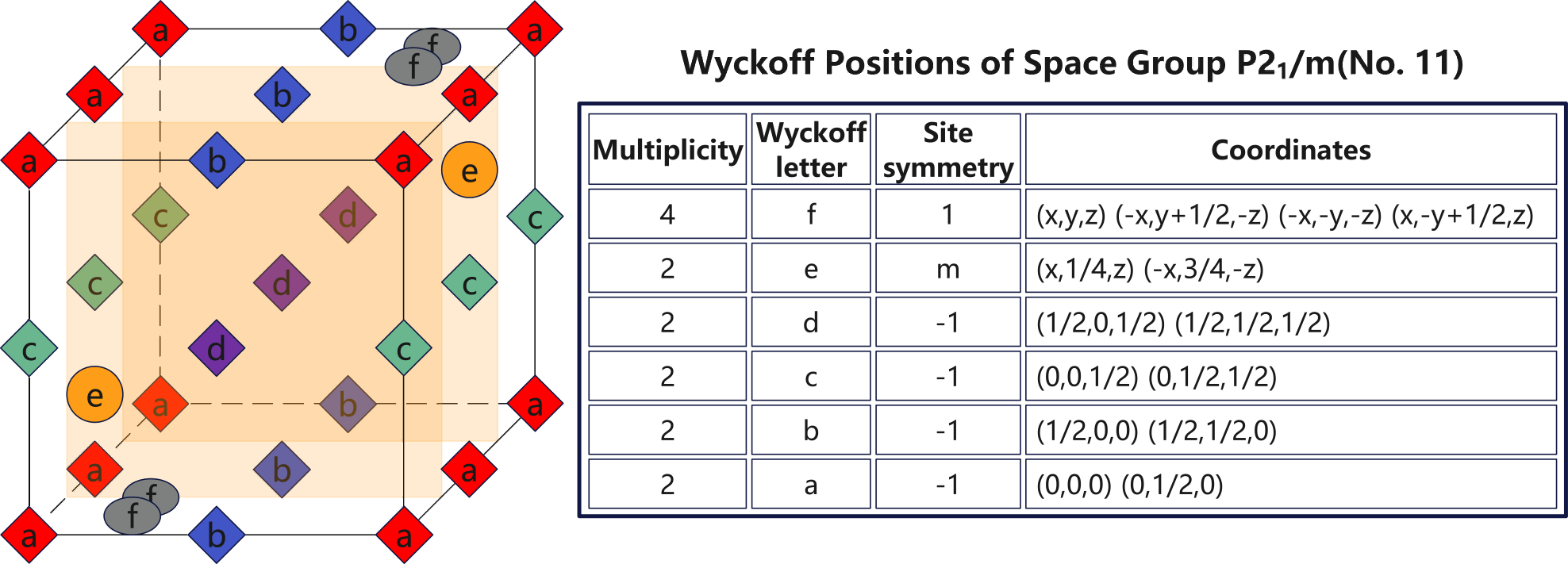}
  \caption{The Impact of Wyckoff Positions of Group $P2_1/m$ (No. 11) on Atomic Coordinates.}
  \label{wp2coord}
\end{figure}

We present a three-dimensional schematic in Figure \ref{wp2coord} to demonstrate the constraints of Wyckoff positions in space group $P2_1/m$ (No. 11) on atomic fractional coordinates. Rhombuses denote Wyckoff letters with fixed coordinates. Due to the periodicity of the unit cell, corner atoms exhibit 1/8 occupancy, edge atoms 1/4 occupancy, and face-centered atoms 1/2 occupancy. Taking Wyckoff letter 'a' as an example, its multiplicity of 2 arises from eight corner-located atoms and four edge-located atoms. Circular and elliptical symbols represent Wyckoff positions with variable coordinates: atoms in circular sites (orange) are constrained by mirror planes with two degrees of freedom, while elliptical sites allow three degrees of freedom governed by interatomic positional relationships.

\section{Our Method CrystalGF}\label{secB}

we present CrystalGF, a novel framework for crystal materials generation that synthesizes desired crystal structures guided by chemical composition and material properties. 

\subsection{Architecture}\label{secB1}
A crystal generation model conditioned with chemical composition and material properties could be described as
${M} = H(C, P)$,
where $H(.)$ represents a crystal structure generator,
$C = ({c_1}, {c_2}, ..., {c_k})$ 
are the chemical composition constraint in which k components are specified, P is property constraint. $M = (A, X, L)$ represents for crystal structure, where 
$A = [{a_1}, {a_2}, ..., {a_n}]$ 
are the elemental composition of atoms inside a unit cell, 
$X = [{x_1}, {x_2}, ..., {x_n}] \in {{[0,1)}^{3 \times n}}$ 
are fractional coordinates of $n$ atoms,
$L = [{l_1}, {l_2}, {l_2}] \in {\mathbb{R} ^ {3 \times 3}}$ 
is the lattice. In this work, we formulate our model as:

$$
\begin{array}{c} 
{M} = H( G(C, P), P) 
\end{array}
$$

where $A, S = G(C, P)$ represents a constraint generator, with $S$ being the symmetry constraint composed of the space group type and Wyckoff positions of the target crystal structure. 
The framework of our two-step model is shown in Figure \ref{CrystalGF}. The constraint generator G(.) takes chemical composition and material properties as input and generates space group, Wyckoff positions and composition ratio, which will be taken as the input of the crystal structure generator H(.) along with material properties.

Notice that for our model, the actual input constraints are still chemical composition and material properties, and we utilize a constraint generator to obtain additional constraints. We utilize a LLMs-based model as our constraint generation module (Appendix \ref{secB2}), and a multi-conditional constrained diffusion model as the crystal structure generation module (Appendix \ref{secB3}). Furthermore, both H(.) and G(.) could be any model as long as the inputs and outputs match.

\begin{figure*}[ht]
  \centering
  \includegraphics[width=1\textwidth]{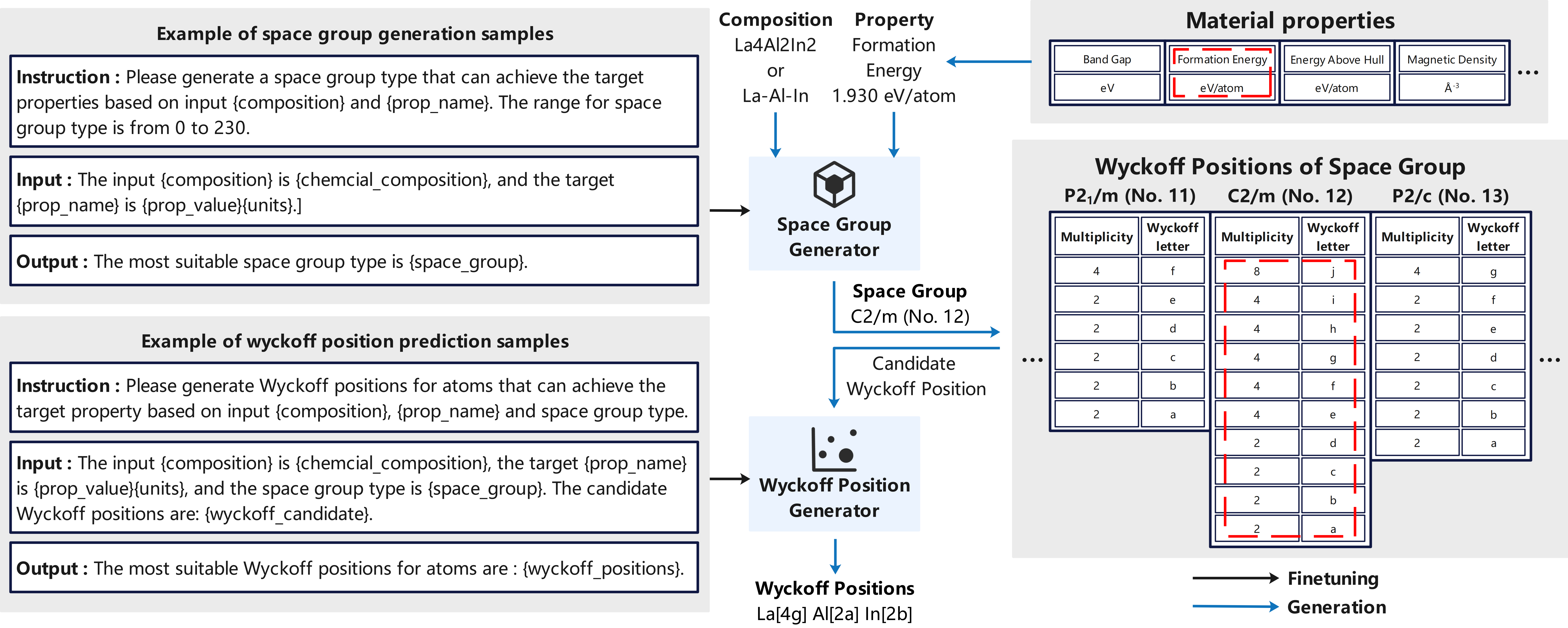}
  \caption{Overview of our constraint generator G(.). The content within curly braces represents the custom model inputs. All our prompts are in Chinese. For review purposes, only the English-translated versions are retained in the figure.}
  \label{CrystalGF-L}
\end{figure*} 

\subsection{Constraint Generator}\label{secB2}

The constraint generator aims to generate constraints for crystal symmetry and composition ratio, where symmetry constraints include space group types and Wyckoff positions. The overall fine-tuning and inference workflow of the module is illustrated in Figure \ref{CrystalGF-L}. Our generator is capable of receiving diverse inputs from chemical  and material properties to structures, where chemical composition can be specified as either elemental composition or full chemical formulas, and material properties could include band gap, formation energy, etc. After embedding these inputs into predefined prompt templates, the module utilizes LLMs to generate corresponding symmetry constraints. Notably, when the chemical composition is provided as elemental composition, the generator simultaneously produces the corresponding composition ratio. 

Since Wyckoff positions generation is related to space group types, we employ two specialized LLMs as space group generator and Wyckoff position generator, to separately generate symmetry information. First, the space group generator determines the space group of crystal unit cell based on input chemical composition and target properties. Subsequently, candidate Wyckoff letters are derived according to the space group index. Finally, the Wyckoff position generator produces corresponding Wyckoff letters for each atom based on the target properties. 

We design the following three tricks in the prompts to regulate and enhance the generation results. (1) \textbf{Generation scope constraints}: For space groups, we explicitly specify in the "Instruction" part that valid space group types should range from 1 to 230; For Wyckoff positions, we provide candidate Wyckoff letters in the "Input" part. (2) \textbf{Numerical standardization}: To minimize the impact of text length on LLMs, we standardized all input material properties by retaining three decimal places and adaptively determining appropriate units based on property types. (3) \textbf{Text optimization}: Instead of directly generating target space group type and Wyckoff positions, we incorporate the phrase "the most suitable..." into the prompts, which has been empirically shown to improve generation accuracy.

\subsection{Crystal Structure Generator}\label{secB3}

The crystal structure generator is designed to receive symmetry constraints, elemental composition, and target properties generated by constraint generator, subsequently generating desired crystal structures. We employ a diffusion model to jointly generate the lattice matrix $\bm{L}$ and fractional coordinates $\bm{F}$ under these conditions. Detailed descriptions of the forward diffusion process and reverse generation procedure are provided in Appendix \ref{C1}.

\begin{figure}[ht]
    \vspace{10pt}
    \centering
    \includegraphics[height=5.2cm]{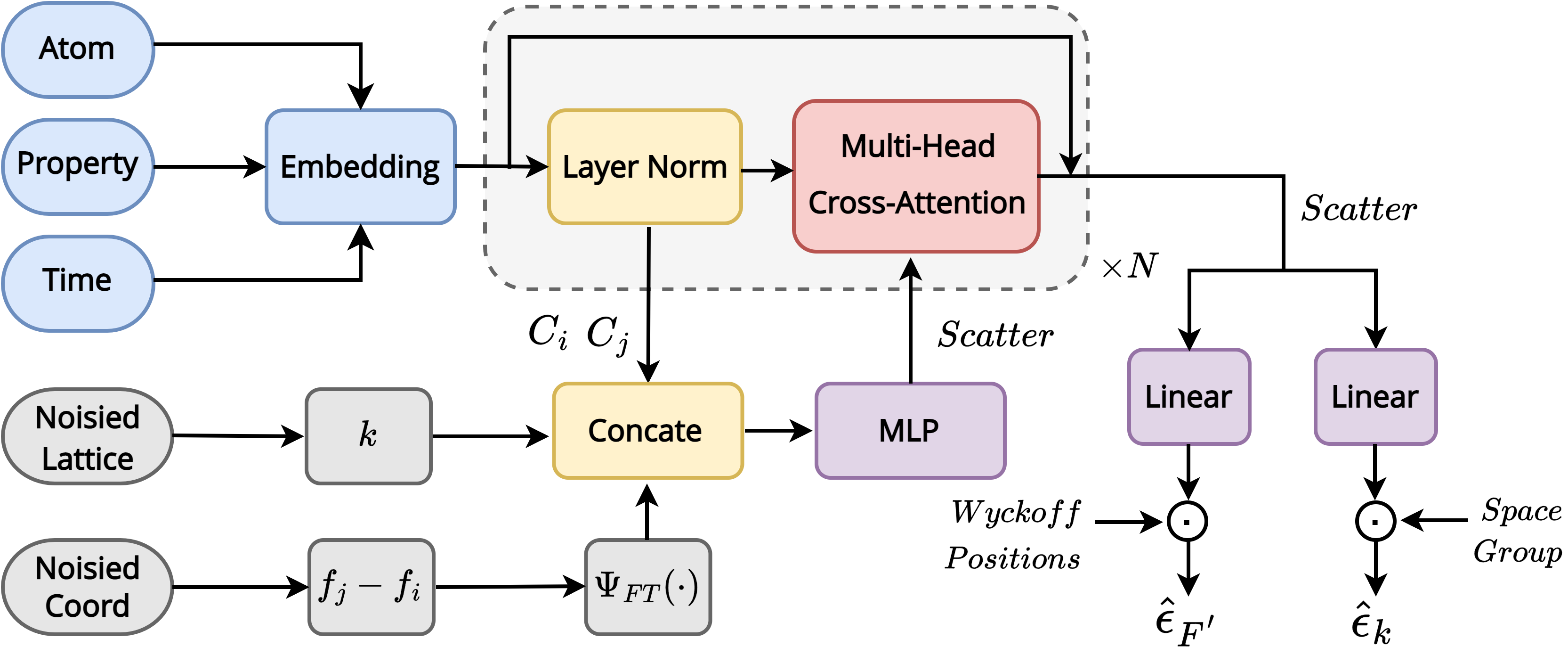}
    \caption{~Diffusion model of crystal structure generator for crystal generation.}\label{CrystalGF-D}
\end{figure}

The overall architecture of our denoising model is illustrated in Figure \ref{CrystalGF-D}, which derives two denoising terms $\hat{\epsilon}_{k}$ , $\hat{\epsilon}_{{F}^{'}} $ under the constraints of elemental composition, crystal symmetry, and target properties. For feature embedding, we introduce continuous material property embeddings into input feature representations. Continuous properties, following Con-CDVAE~\cite{ye2024cdvae}, are expanded by a Gaussian Radial Basis Function (RBF) and embedded by an MLP:

\begin{equation}\label{prop} 
    f_{prop} \left( p \right)  = \\ 
    \varphi_{mlp} \left( \left[ e^{\frac{-(p-p_{min})^2}{2\sigma^2}} \text{,} e^{\frac{-(p-p_{min}-\sigma)^2}{2\sigma^2}},...,e^{\frac{-(p-p_{max})^2}{2\sigma^2}}\right] \right)
\end{equation}

\noindent where $p$ is property value, $p_{min}$ is the preset minimum property value, $p_{max}$ is the maximum property value, and $\sigma$ is grid spacing of RBF. We concatenate the atom embeddings $f_{atom}\left( A \right)$, property embeddings $f_{prop}\left( p \right)$ and the sinusoidal time embedding $f_{time}\left( t \right)$ to get the input features $C = \phi_{in} \left( f_{atom}\left( A \right) \text{,} f_{prop}\left( p \right) \text{,}  f_{time}\left( t \right) \right)$. where $\phi_{in}$ is a linear layer. The embedding of material properties enriches the input feature information. Compared to the strict symmetry constraints in high space groups, this approach allows the model to prioritize property-driven generation of anticipated crystal structures when operating in lower space groups with higher degrees of freedom.

For the embedding of noised lattices and noised coordinates, we adopt the approach described in DiffCSP++~\cite{jiao2024space}. Subsequently, we employ a multi-level cascade decoder to generate denoising terms. To enhance the model's fitting capacity, we incorporate a multi-head cross-attention mechanism within the decoder architecture to fuse information from both the noised feature and the input feature. The structure of the decoder in the n-th layer is designed as 

\begin{equation}
\label{layer}
    C^{\left( n \right) }_{ij} = \varphi_e \left( C^{\left( n-1 \right) }_i,C^{\left( n-1 \right) }_j,k,\psi_{FT} \left( f_j-f_i \right) \right),
\end{equation}

\begin{equation}\label{mhca}
    C^{\left( n \right) } = C^{\left( n-1 \right) } + \varphi_{mhca}(C^{\left( n-1 \right) },\sum_{j=1}^{N}C^{\left( n \right) }_{ij}),
\end{equation}


\noindent where in Eq.(\ref{layer}), $ C^{\left( n-1 \right) }_i $ and $ C^{\left( n-1 \right) }_j$ are node features extracted from $ C^{\left( n-1 \right) } $, $k$ is the unique O(3)-invariant representation of the lattice, and $\psi_{FT}:(-1,1)^3 \to [-1,1]^{3 \times K}$ is the Fourier transformation with $K$ bases to periodically embed the relative fractional coordinate $f_j - f_i$. In Eq.(\ref{mhca}) $\varphi_{mhca}$ is multi-head cross-attention to fuse the noised features and coordinates and the input features. 

\section{Implementation Details}\label{secC}

\subsection{Hyper-parameters and Training Details}\label{C1}

We train CDVAE~\cite{xie2021crystal}, DiffCSP++~\cite{jiao2024space}, and our crystal structure generator on three public datasets to validate the effectiveness of the structure generator. Using the dataset described in Section \ref{subsec2.1}, we train CDVAE~\cite{xie2021crystal}, DiffCSP++~\cite{jiao2024space}, FlowMM~\cite{miller2024flowmm}, SymmCD~\cite{levy2025symmcd}, Con-CDVAE~\cite{ye2024cdvae}, MatterGen~\cite{zeni2023mattergen}, and our CrystalGF to verify the validity of the overall framework. We split our data into 80\% training, 10\% validation, and 10\% test sets while maintaining strict correspondence between space group and Wyckoff positions entries, this opeartion prevents data leakage by ensuring identical crystals are allocated to the same subset. To the best of our knowledge, Con-CDVAE and MatterGen represent the advanced multi-conditional generative models, which utilize chemical formulas with material properties and elemental composition with material properties as model inputs, respectively. Therefore, we conduct comparative evaluations of both models against our CrystalGF.

For CDVAE, we apply the transformer-enhanced~\cite{vaswani2017attention} version of CDVAE as introduced in DiffCSP~\cite{jiao2024crystal}, while maintaining consistent training parameters. For DiffCSP++, we train the denoising model with 6 layers, 512-dimension hidden layer, and 128-dimension Fourier embeddings and the training epochs are set to 3000, 1500, 1000 for Perov-5, MP-20, and MPTS-52. For FlowMM and SymmCD, we trained two models in unconditional mode and ab initio generative mode under default Settings, respectively. For MatterGen, we fine-tuning the pre-trained model with chemical system and property, including band gap and formation energy. The dimension of the property embedding and the chemical system multi-hot embedding is set to 512, and model is trained for 1000 epochs with an Adam optimizer with initial learning rate $5 \times 10^{-6}$ and a Plateau scheduler with a decaying factor 0.6 and a patience of 100 epochs. For Con-CDVAE, we train model with chemical formula and property, including band gap and formation energy. The formula embedding module has 3 layers, 92-dimension input layer and 64-dimension hidden layer, the property embedding module has 3 layers and 64-dimension hidden layer and the model is trained for 1000 epochs with an Adam optimizer with initial learning rate $1 \times 10^{-3}$ and a Plateau scheduler with a decaying factor 0.6 and a patience of 30 epochs.

For our symmetry generator, We fine-tune four LLMs ,including Qwen2.5-7B-Instruct, Llama-3.1-8B-Instruct, DeepSeek-R1-8B, and GLM-4-9B-chat, using LoRA from Llamafactory~\cite{zheng2024llamafactory} framework. The LoRA fine-tuning specifically targets the qkv layers in transformer modules with 20 epochs and a $1 \times 10^{-4}$ initial learning rate, the learning rate scheduler type is set to cosine annealing strategy with 0.1 warm up ratio. And the fine-tuning precision is set to FP32 throughout all computations. 

For our Structure generator, we apply the input features consist of 512-dimension atom embeddings, 128-dimension property embeddings and 128-dimension Fourier embeddings. The decoder employs 6 layers with 512 hidden states and multi-head cross-attention using num\_heads = 2. The training epochs are configured as 3000 for Perov-5, 1500 for MP-20, and 1000 for MPTS-52. Optimization utilizes Adam with an initial learning rate of $1 \times 10^{-3}$, coupled with a ReduceLROnPlateau scheduler that reduces the learning rate by 0.6 factor and 30 patience. All models are trained on A40 GPU.

\subsection{Symmetry Accuracy Details}\label{C2}

For the evaluation of space group generation accuracy, we directly compare the outputs with the labels from the dataset. Regarding Wyckoff position generation accuracy, since different Wyckoff letters correspond to varying multiplicities across space groups, we exclusively calculate the accuracy of Wyckoff letters. Notably, a type of atom may possess multiple Wyckoff letters, and the generation order of these letters should not affect accuracy statistics. For instance, the sequences Ba[4d] Ba[6g] and Ba[6g] Ba[4d] should be considered equivalent. 

When constructing the confusion matrix for Wyckoff positions generation, we first identify matches where both the atomic order and generated letters coincide. Second, we distinguish cases where generated letters match despite atomic order discrepancies. Finally, atoms with inconsistent generation quantities are handled by padding with an "other" category.

\subsection{DFT Calculation Details}\label{C3}

We employ density functional theory (DFT) for computational simulations using the VASP~\cite{kresse1996efficient}~\cite{kresse1996efficiency},which implements the projector augmented wave (PAW) method, to conduct material property simulations and validation calculations. The generalized gradient approximation (GGA) with the Perdew-Burke-Ernzerhof (PBE) functional was adopted to treat exchange-correlation interactions. In our simulations, we set a plane-wave energy cutoff of 520 eV for the basis set and utilized a Gamma-centered k-point mesh with a grid spacing of 0.06$\pi$ $\angstrom^{-1}$. The total energy convergence criterion was set to 1.0 $\times$ $10^{-1}$ eV. Post-processing of data obtained from VASP was performed using the VASPKIT toolkit.

We calculate band gap and formation energy of generated materials. The band gap is defined as the energy interval between the valence band maximum (VBM) and the conduction band minimum (CBM). Using VASP, we performed two-step calculations : self-consistent field (SCF) and non-self-consistent field (NSCF) computations, followed by post-processing with vaspkit to obtain simulation-validated bandgap values, which were compared with generated band gap results.

The formation energy refers to the energy released when a compound is synthesized from its constituent elemental phases. For a binary compound $A_mB_n$, its formation energy can be expressed as:

\begin{equation}
    E_f=\frac {E \left( A_m B_n \right)-m \times E \left( A \right) - n \times E \left( B \right)}{ m + n }
    \label{fm}
\end{equation}

\noindent where $E \left( A_m B_n \right)$ represents the total energy of the compound $A_mB_n$, while $E \left( A \right)$ and $E \left( B \right)$ denote the normalized energies of the corresponding elements A and B, respectively. 

By using a batch processing method, we calculated the total energy of the compound and the corresponding elements using the VASP software package, and obtained the formation energy based on the Eq.(\ref{fm}).

\begin{table}[t]
  \caption{The constraint of space group types on lattice shape, where a, b, c represent the lengths of the lattice vectors, and $\alpha$, $\beta$, $\gamma$ represent the angles between them.}
  \label{compute resources}
  \centering
  \begin{tabular}{cccc}
    \toprule
    Model Training & Computer Work & Memory Usage & Time Usage\\
    \midrule
    \makecell{CDVAE \\ Con-CDVAE \\ DiffCSP++ \\ MatterGen \\ Constraint Generator \\ Structure Generator } & NVIDIA A40 GPU & \makecell{12GB \\ 24GB \\ 24GB \\ 45GB \\ 26GB + 30GB \\ 30GB} & \makecell{10h \\ 12h \\ 30h \\ 50h \\ 9h + 18h \\ 46h } \\
    \bottomrule 
  \end{tabular}
\end{table}

\subsection{Experiments Compute Resources}\label{C4}

We present the computational resources consumed for training in Table \ref{compute resources}, including GPU type, memory usage, and training duration. The resource consumption for CDVAE and DiffCSP++ corresponds to training with the MP-20 dataset mentioned in Appendix \ref{subsec2.2}, while Con-CDVAE, MatterGen, the Constraint Generator, and the Structure Generator reflect the resource usage when trained on the dataset described in Appendix \ref{subsec2.1}.\\

In the DFT calculations, we perform VASP simulations on the generated crystal structures using the Hygon 7380 CPU. The formation energy calculations required 1.92 core-hours per structure, while the band gap calculations consumed 3.84 core-hours per structure and we consume approximately 100000 core-hours in total.

\section{Visualization of Generated Structures}\label{D}

\begin{figure*}[ht]
	\centering
	\begin{minipage}{0.18\linewidth}
		\centering
            \text{mp-1069603}
            \text{\footnotesize{FE: -0.661 eV/atom}}
		\includegraphics[height=2.1cm]{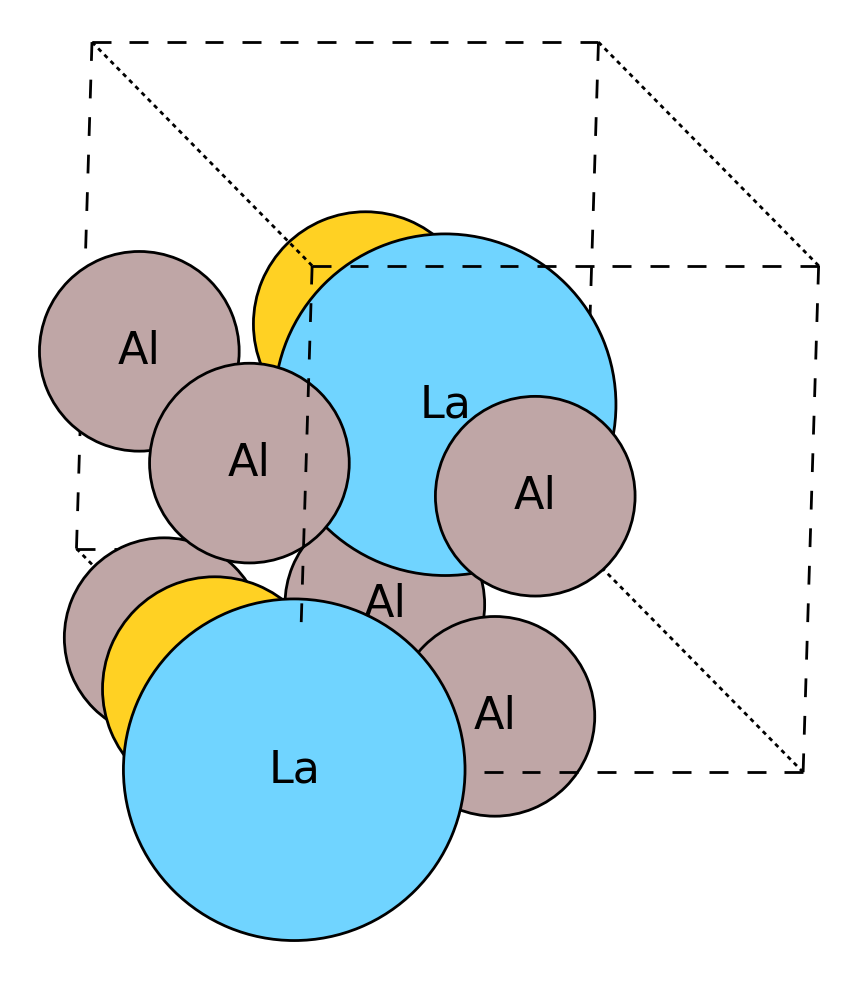}
	\end{minipage}
	\begin{minipage}{0.14\linewidth}
		\centering
            \text{mp-1029574}
            \text{\footnotesize{BG: 0.000 eV}}
		\includegraphics[height=2.1cm]{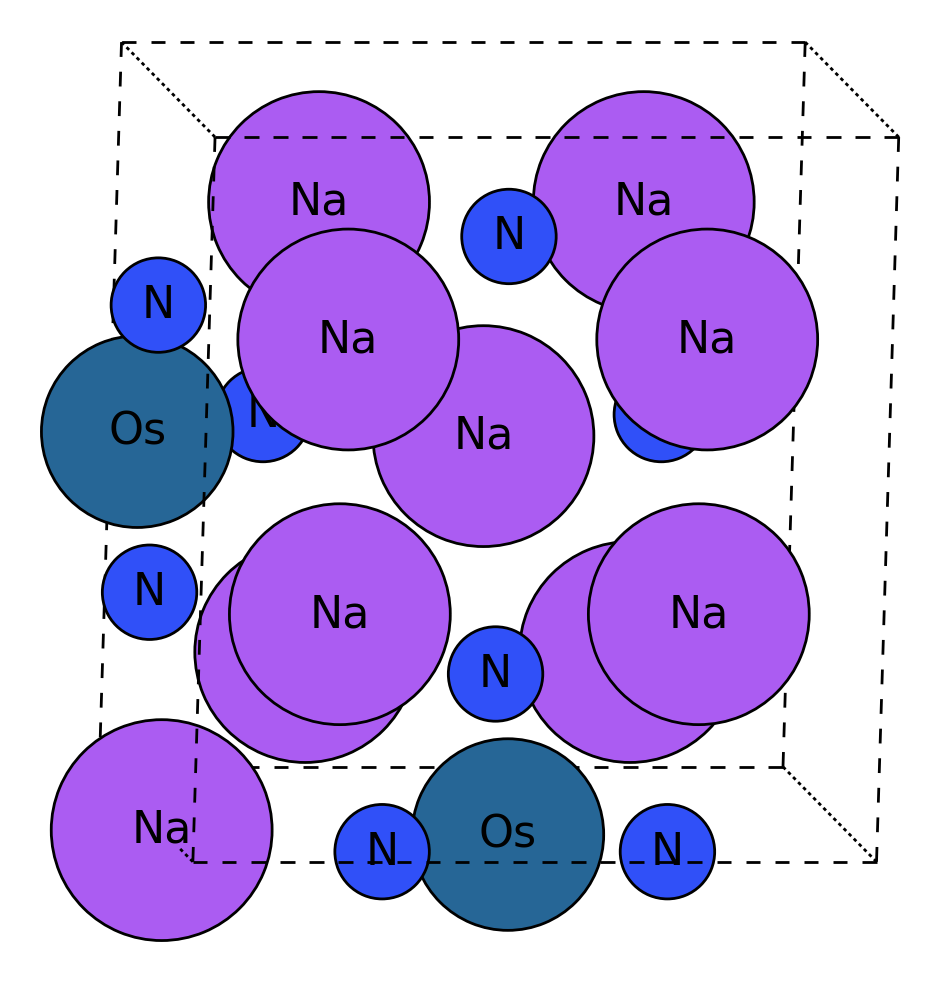}
	\end{minipage}
	\begin{minipage}{0.18\linewidth}
		\centering
            \text{mp-1253073}
            \text{\footnotesize{FE: -3.186 eV/atom}}
		\includegraphics[height=2.1cm]{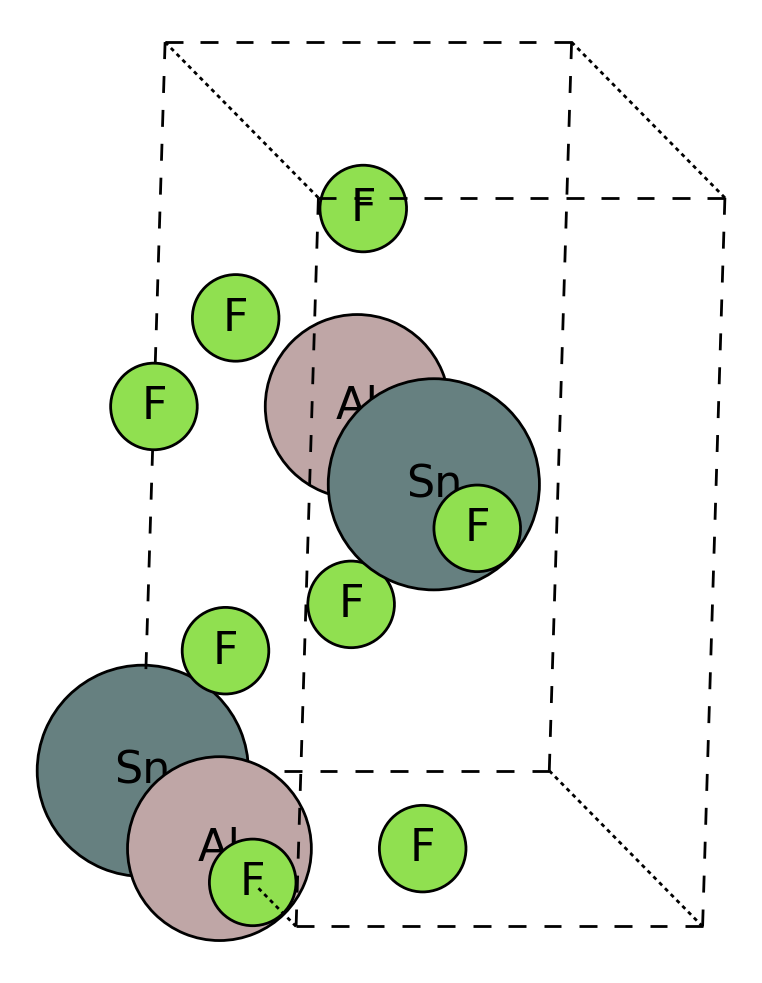}
	\end{minipage}
        \begin{minipage}{0.14\linewidth}
		\centering
            \text{mp-8196}
            \text{\footnotesize{BG: 1.720 eV}}
		\includegraphics[height=2.1cm]{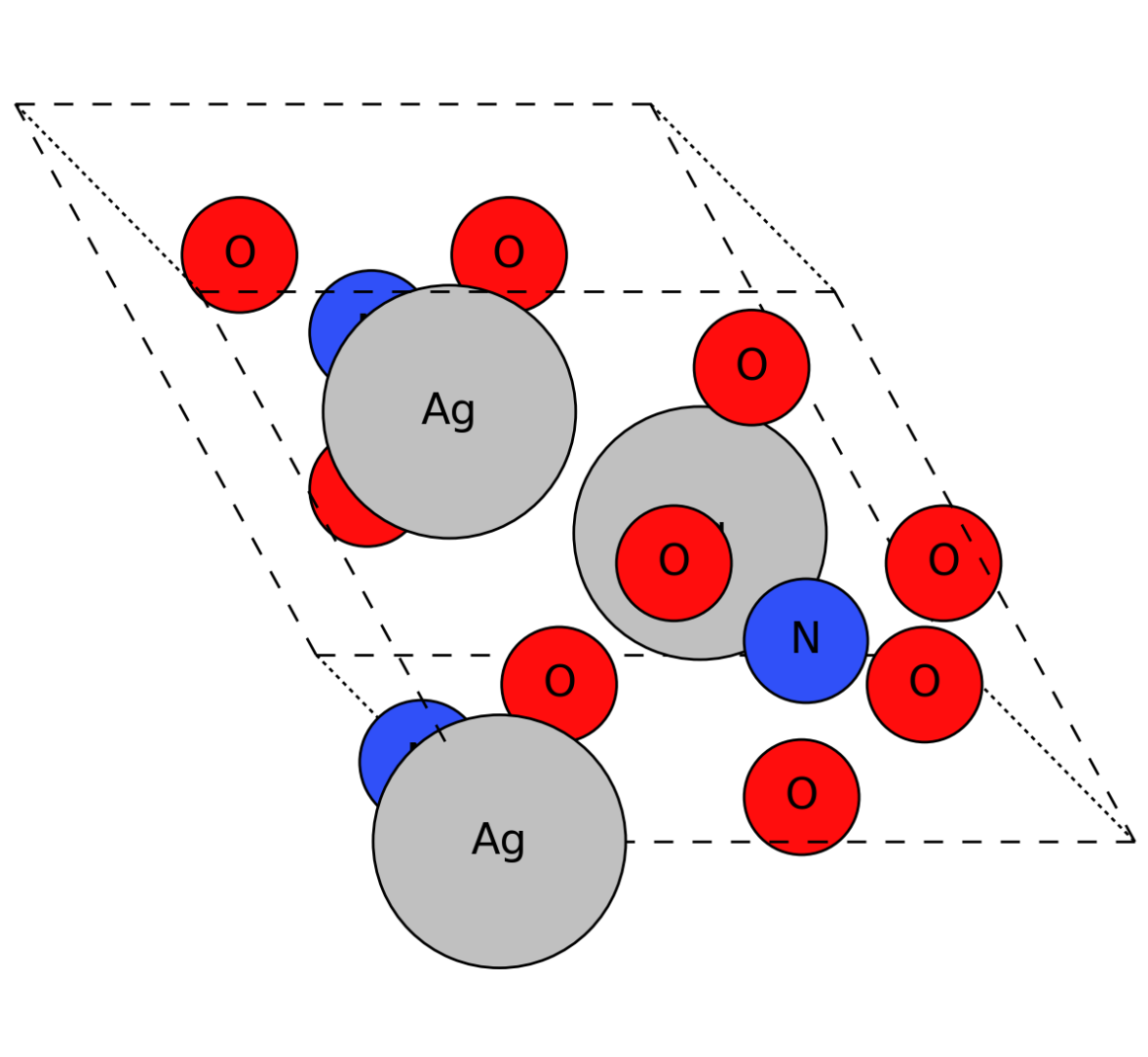}
	\end{minipage}
	\begin{minipage}{0.18\linewidth}
		\centering
            \text{mp-1183716}
            \text{\footnotesize{FE: -0.118 eV/atom}}
		\includegraphics[height=2.1cm]{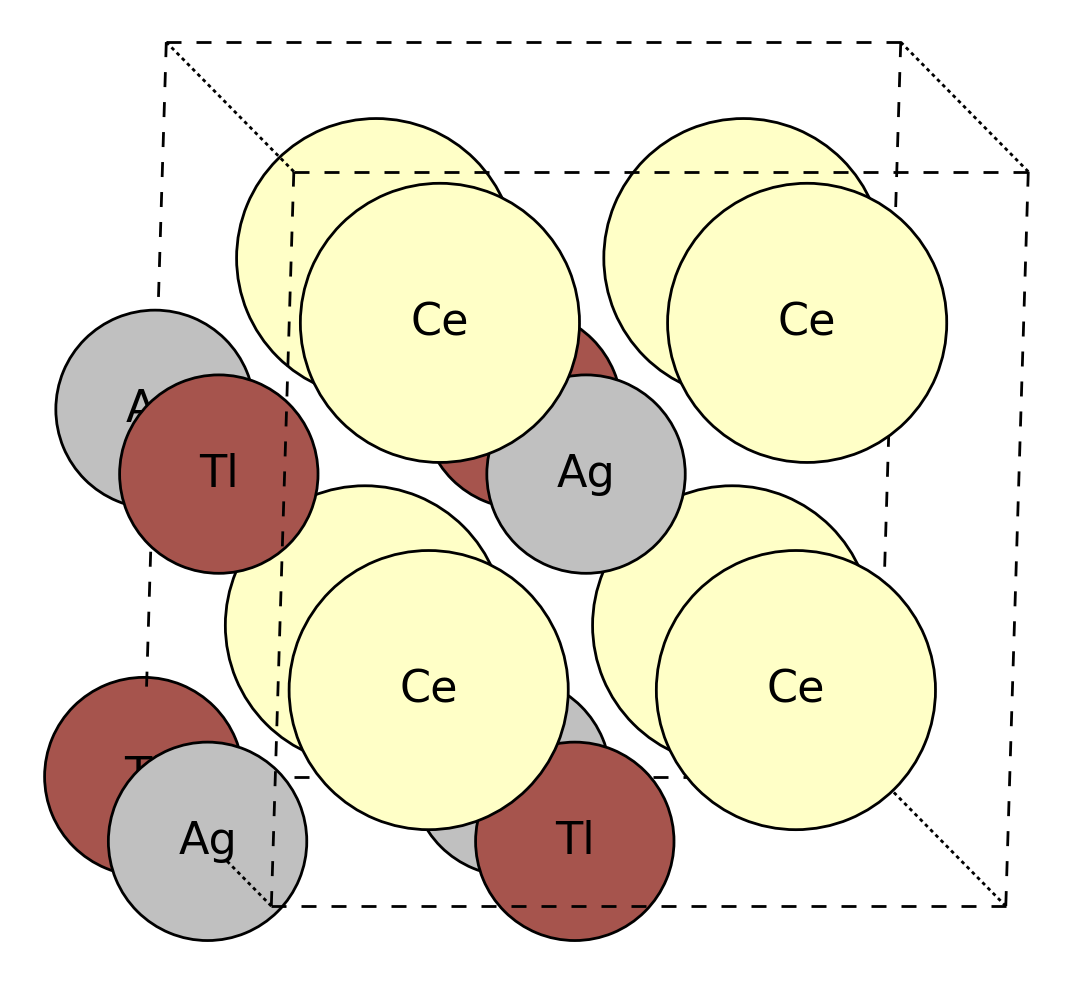}
	\end{minipage}
        \begin{minipage}{0.14\linewidth}
		\centering
            \text{mp-1229121}
            \text{\footnotesize{BG: 0.068 eV}}
		\includegraphics[height=2.1cm]{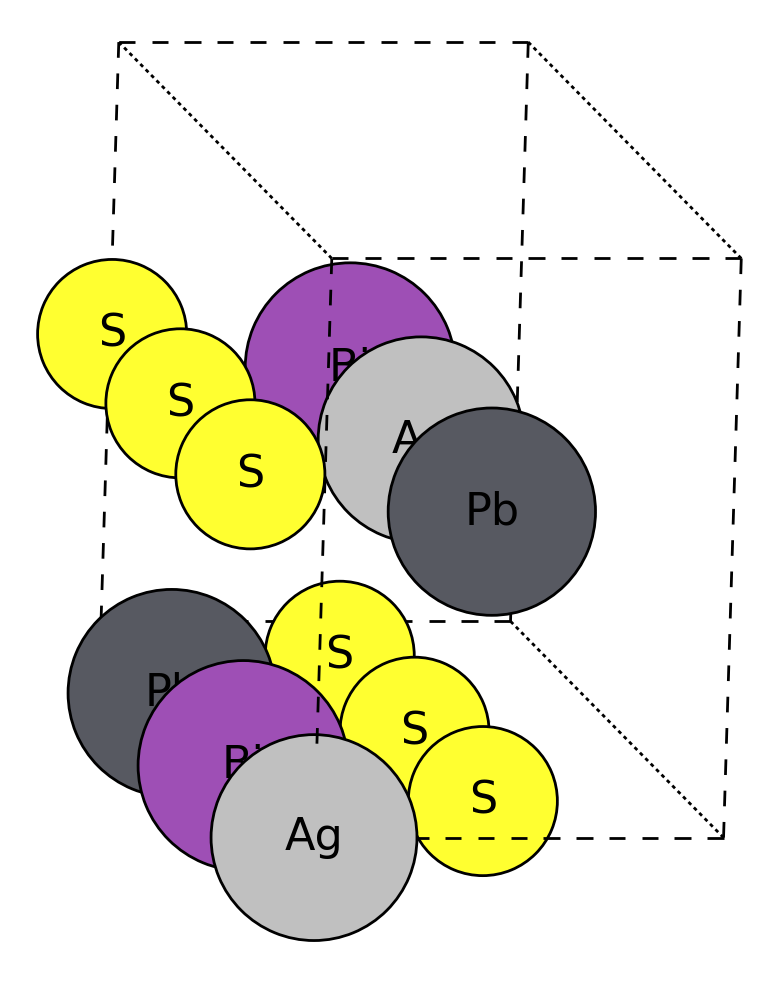}
	\end{minipage}
	
	\centering
	\begin{minipage}{0.18\linewidth}
		\centering
            \text{$\text{La}_{2}\text{Al}_{6}\text{Au}_{2}$}
            \text{\footnotesize{FE: -0.739 eV/atom}}
		\includegraphics[height=2.1cm]{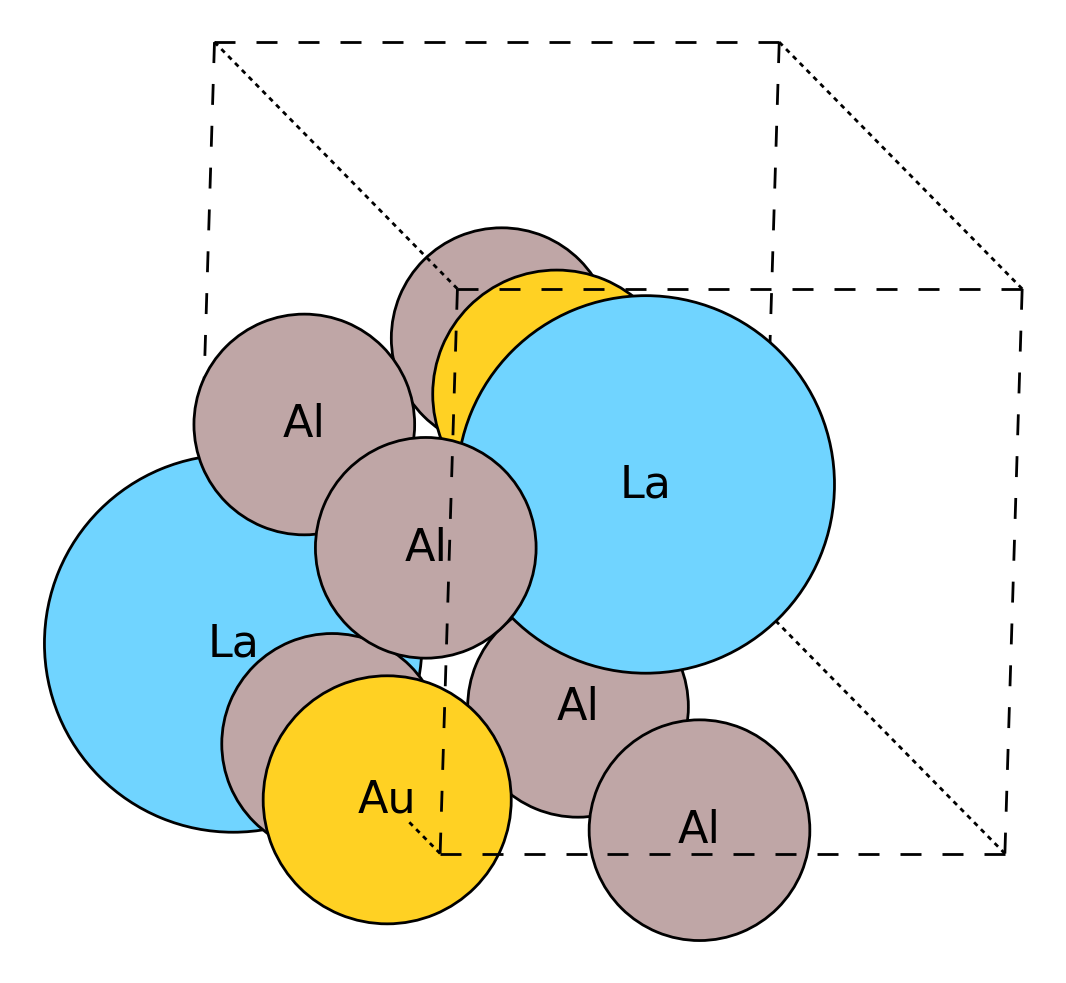}
	\end{minipage}
	\begin{minipage}{0.14\linewidth}
		\centering
            \text{$\text{Na}_{10}\text{Os}_{2}\text{N}_{8}$}
            \text{\footnotesize{BG: 0.000 eV}}
		\includegraphics[height=2.1cm]{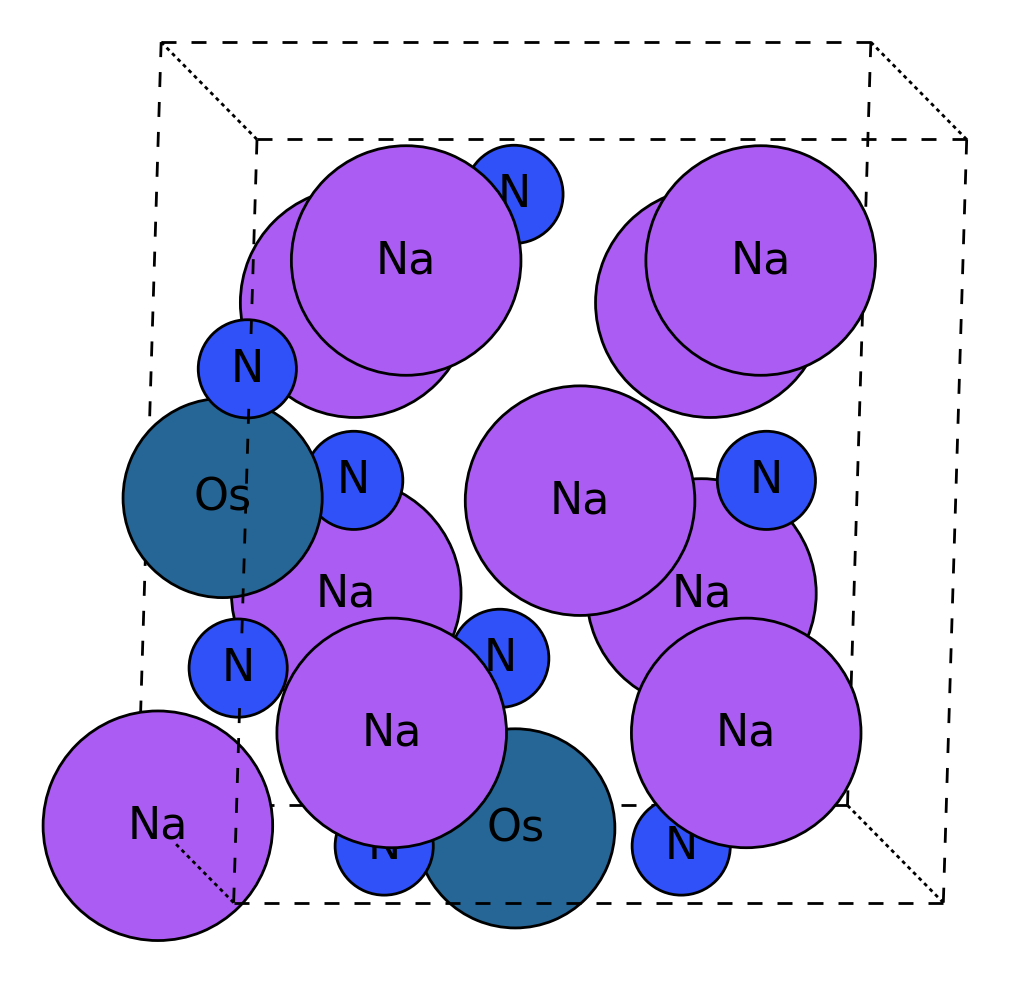}
	\end{minipage}
	\begin{minipage}{0.18\linewidth}
		\centering
            \text{$\text{Al}_{2}\text{Sn}_{2}\text{F}_{10}$}
            \text{\footnotesize{FE: -2.934 eV/atom}}
		\includegraphics[height=2.1cm]{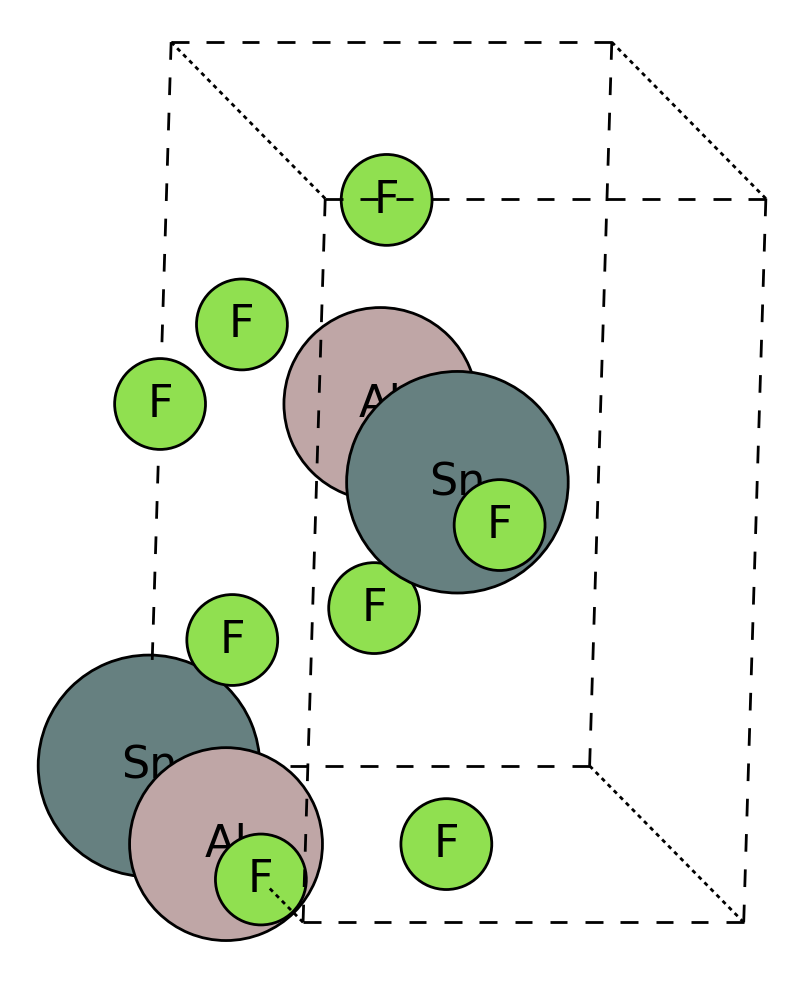}
	\end{minipage}
	\begin{minipage}{0.14\linewidth}
		\centering
            \text{$\text{Ag}_{3}\text{N}_{3}\text{O}_{9}$} 
            \text{\footnotesize{BG: 1.626 eV}}
		\includegraphics[height=2.1cm]{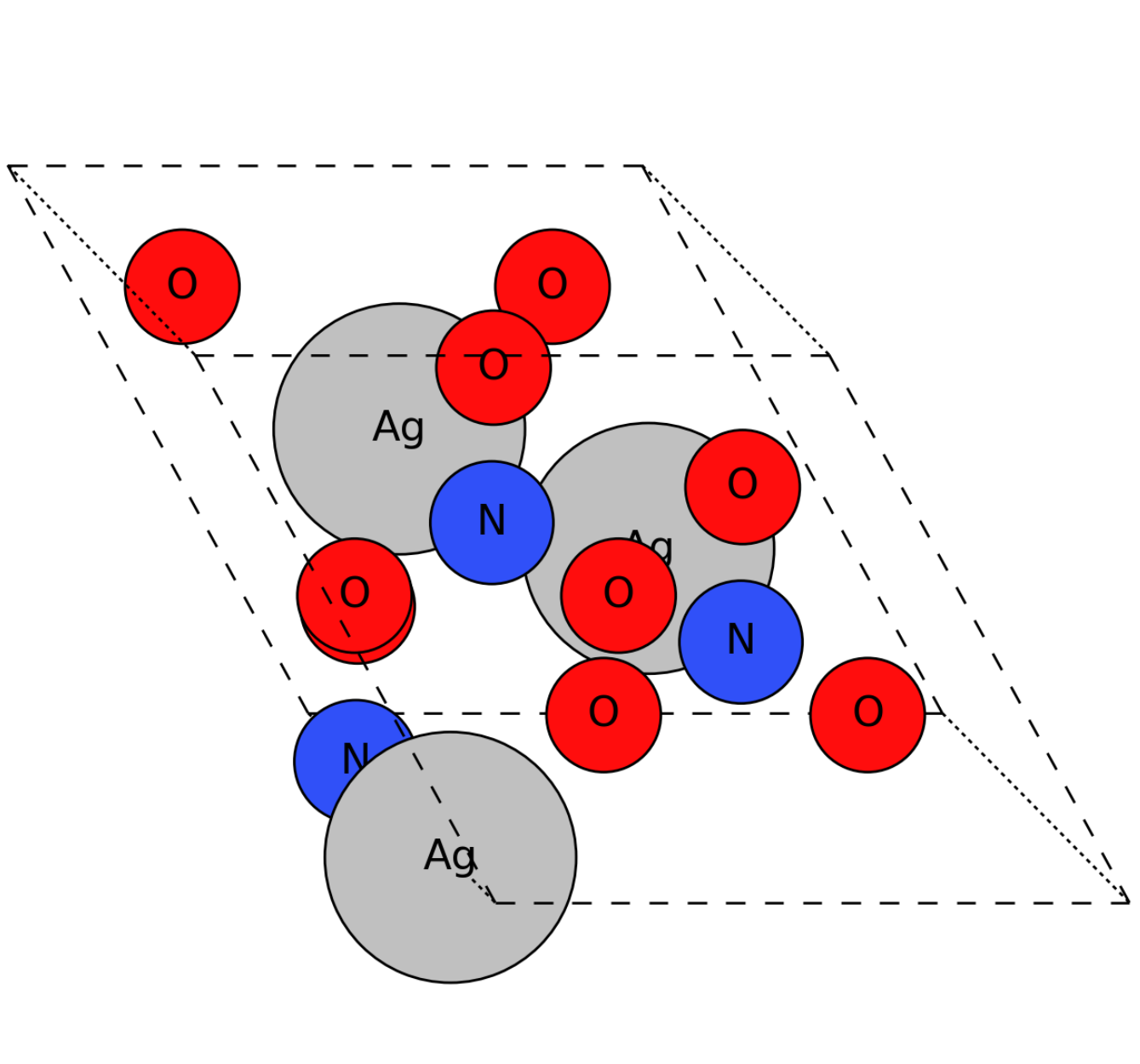}
	\end{minipage}
	\begin{minipage}{0.18\linewidth}
		\centering
            \text{$\text{Ce}_{8}\text{Tl}_{4}\text{Ag}_{4}$}
            \text{\footnotesize{FE: -0.126 eV/atom}}
		\includegraphics[height=2.1cm]{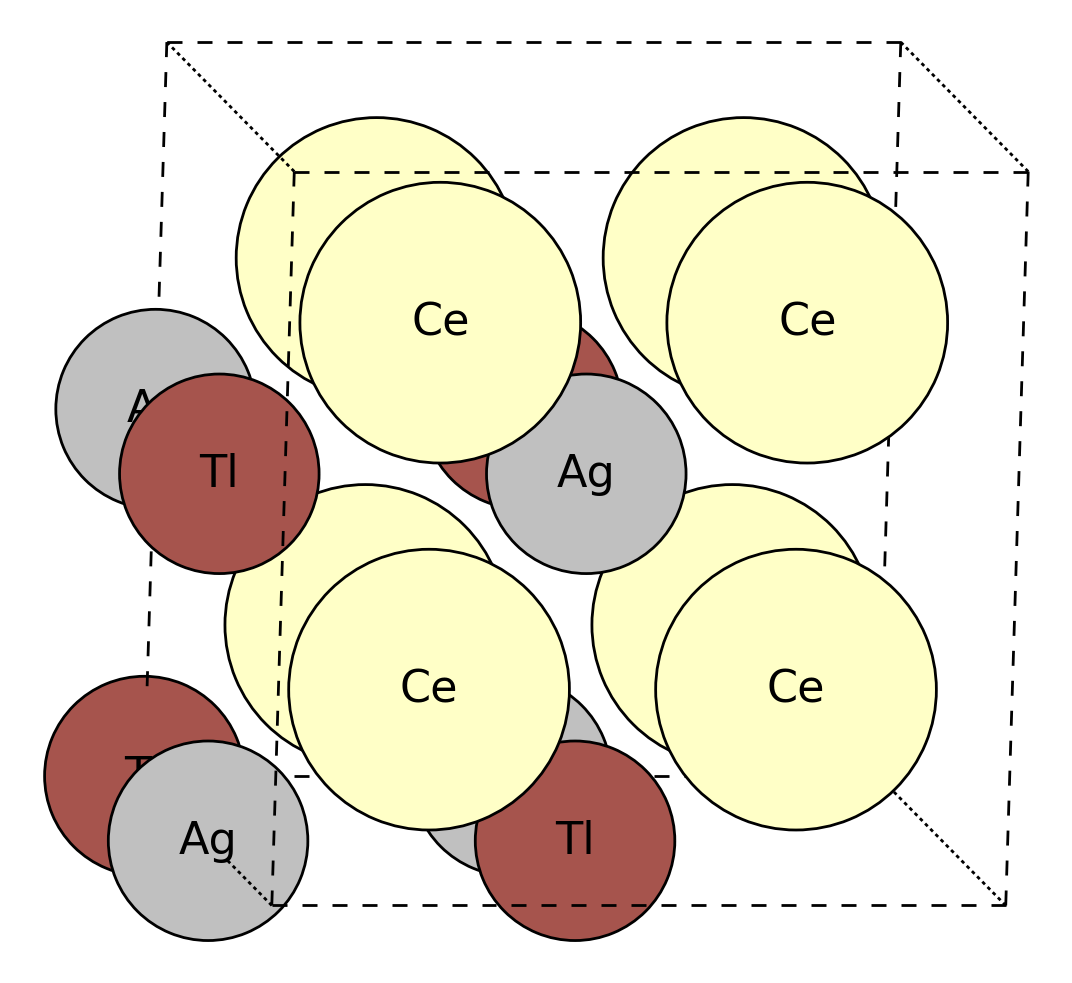}
	\end{minipage}
	\begin{minipage}{0.14\linewidth}
		\centering
            \text{$\text{Ag}_{2}\text{Bi}_{2}\text{Pb}_{2}\text{S}_{6}$} 
            \text{\footnotesize{BG: 0.066 eV}}
		\includegraphics[height=2.1cm]{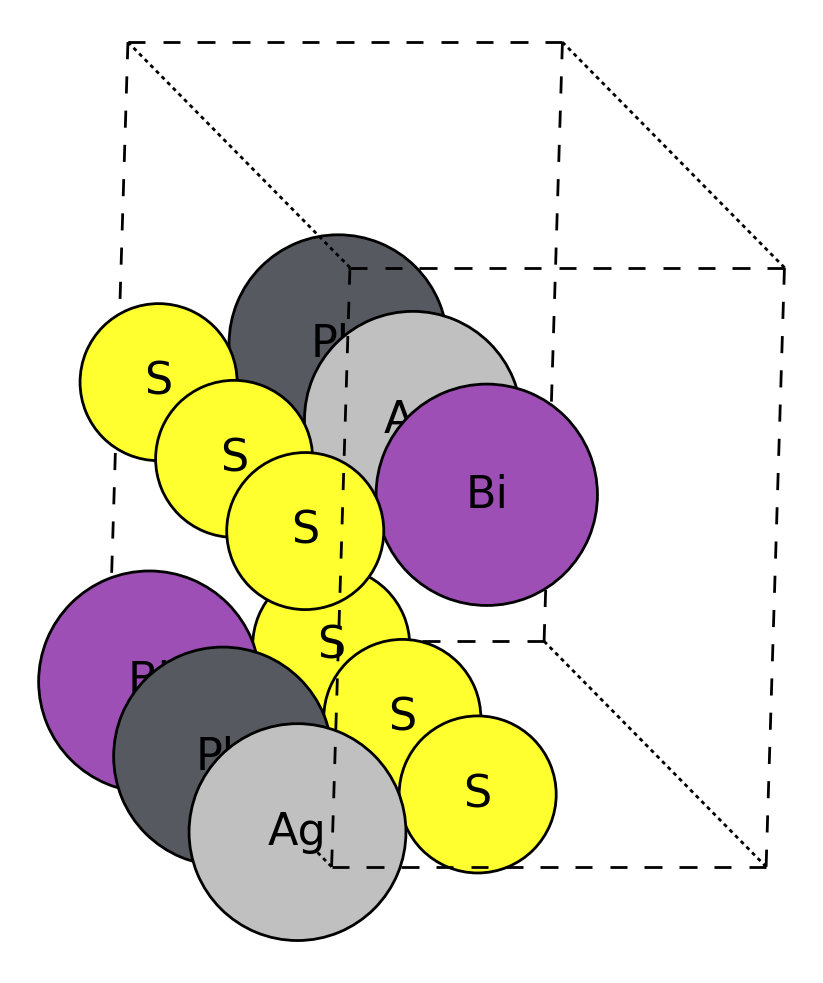}
	\end{minipage}
    \caption{Crystal structures generated by CrystalGF exhibit the same symmetry and similar material properties as those in the MP database.}
    \label{same}
\end{figure*}

\begin{figure*}[ht]
	\centering
	\begin{minipage}{0.18\linewidth}
		\centering
            \text{mp-1080118}
            \text{\footnotesize{FE: -0.505 eV/atom}}
		\includegraphics[height=2.1cm]{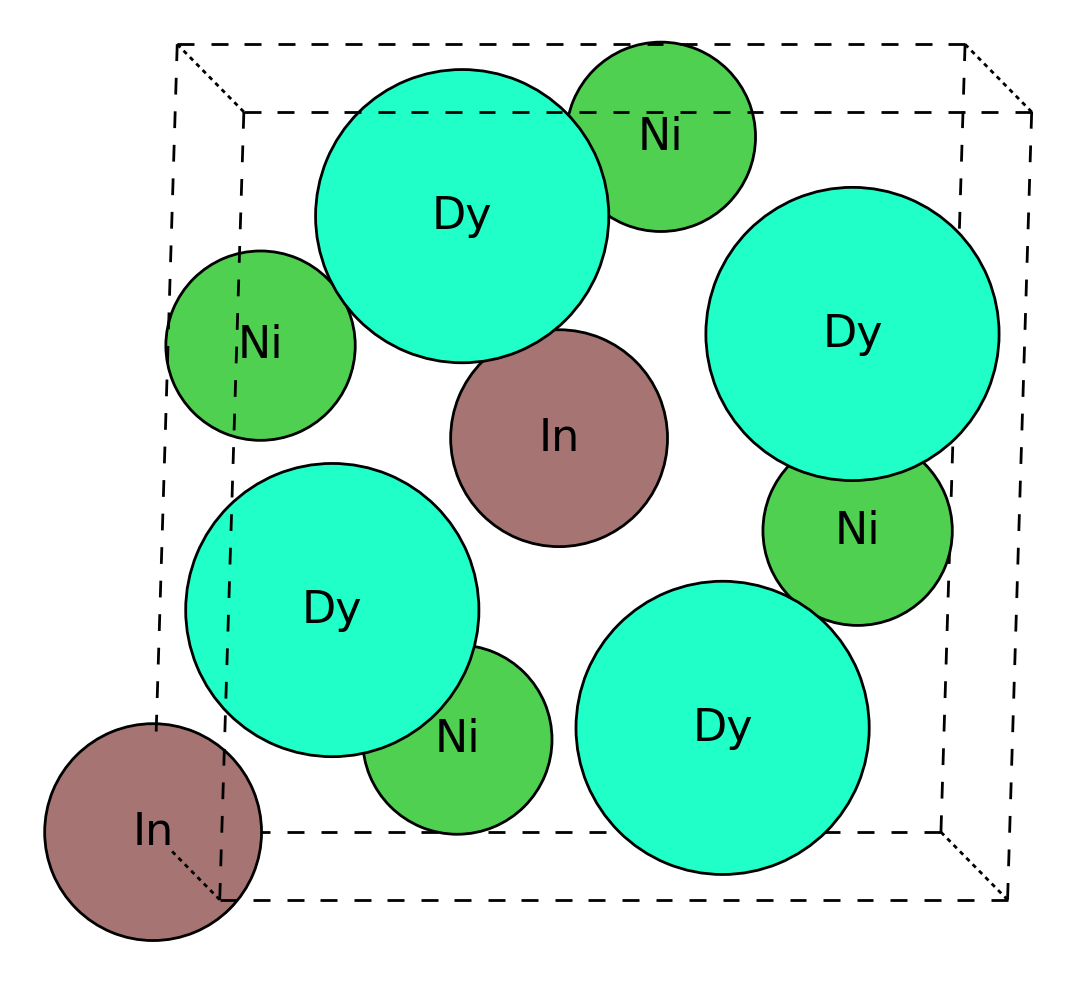}
	\end{minipage}
	\begin{minipage}{0.14\linewidth}
		\centering
            \text{mp-19447}\\
            \text{\footnotesize{BG: 3.703 eV}}
		\includegraphics[height=2.1cm]{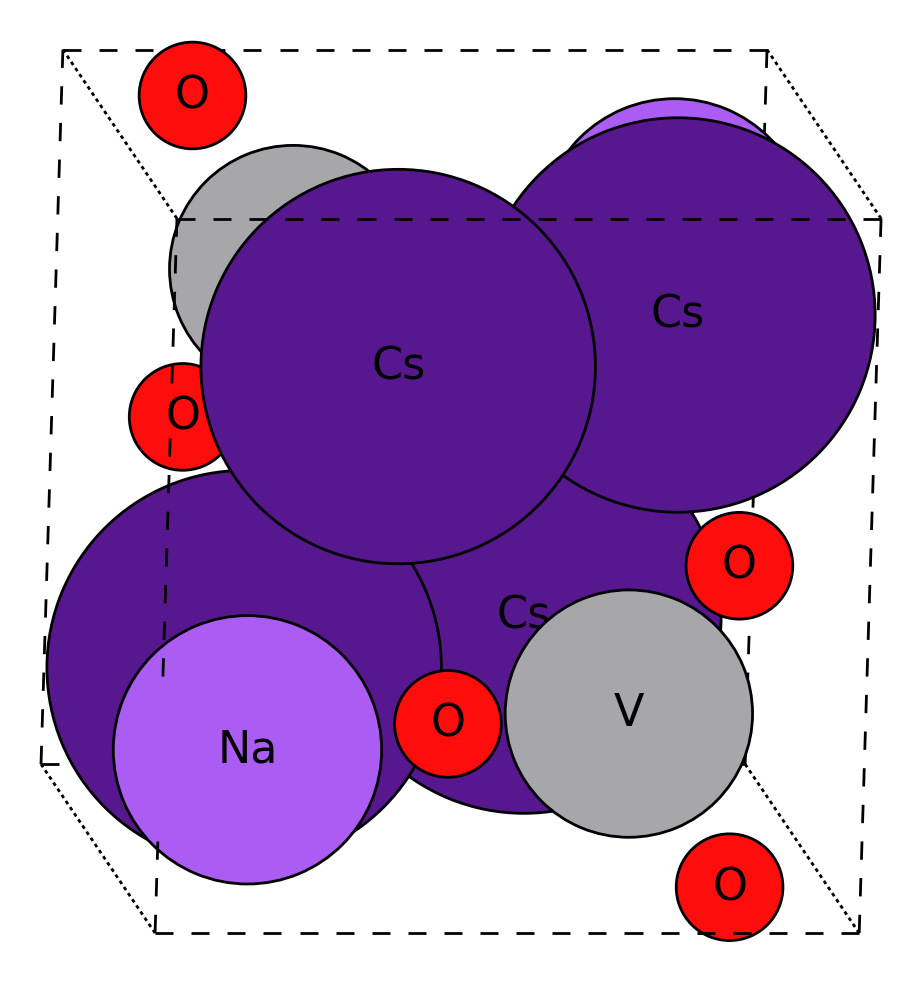}
	\end{minipage}
	\begin{minipage}{0.18\linewidth}
		\centering
            \text{mp-1522247}
            \text{\footnotesize{FE: -2.449 eV/atom}}
		\includegraphics[height=2.1cm]{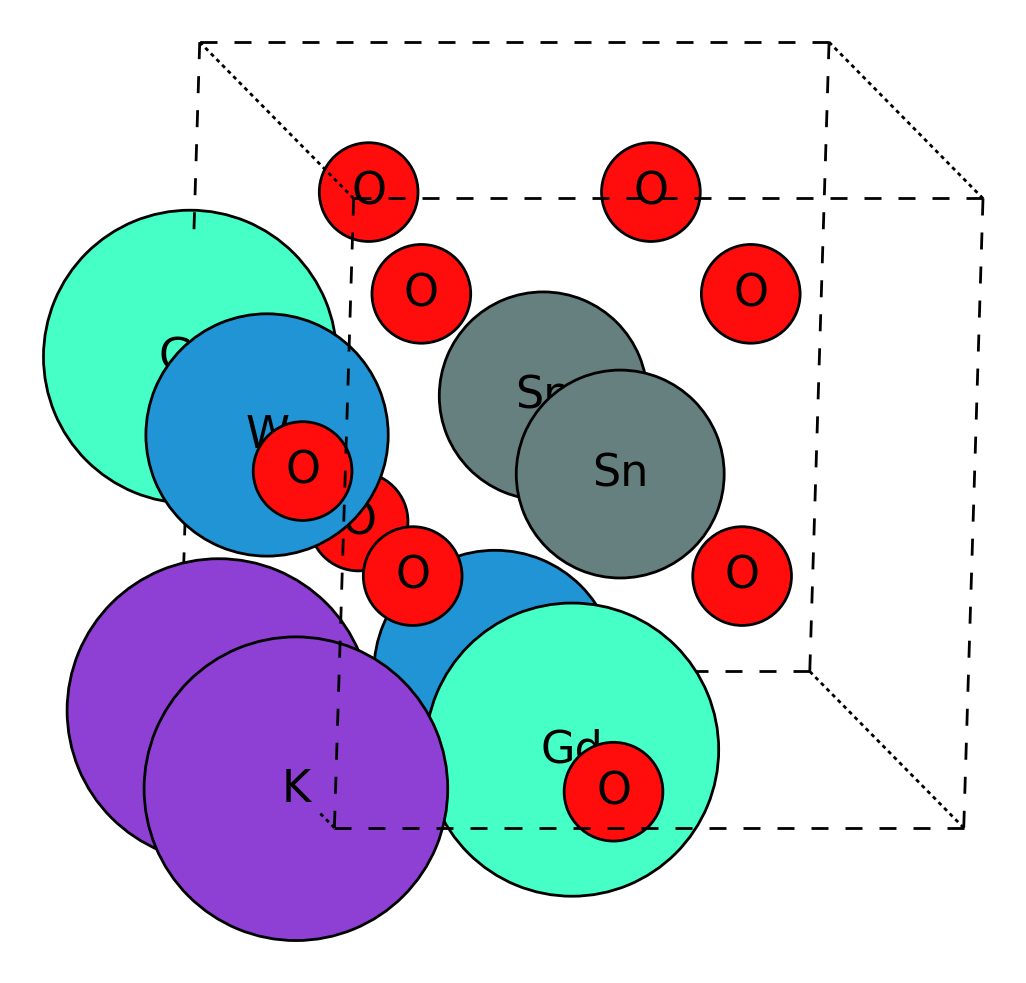}
	\end{minipage}
	\begin{minipage}{0.14\linewidth}
		\centering
            \text{mp-29873}\\
            \text{\footnotesize{BG: 3.168 eV}}
		\includegraphics[height=2.1cm]{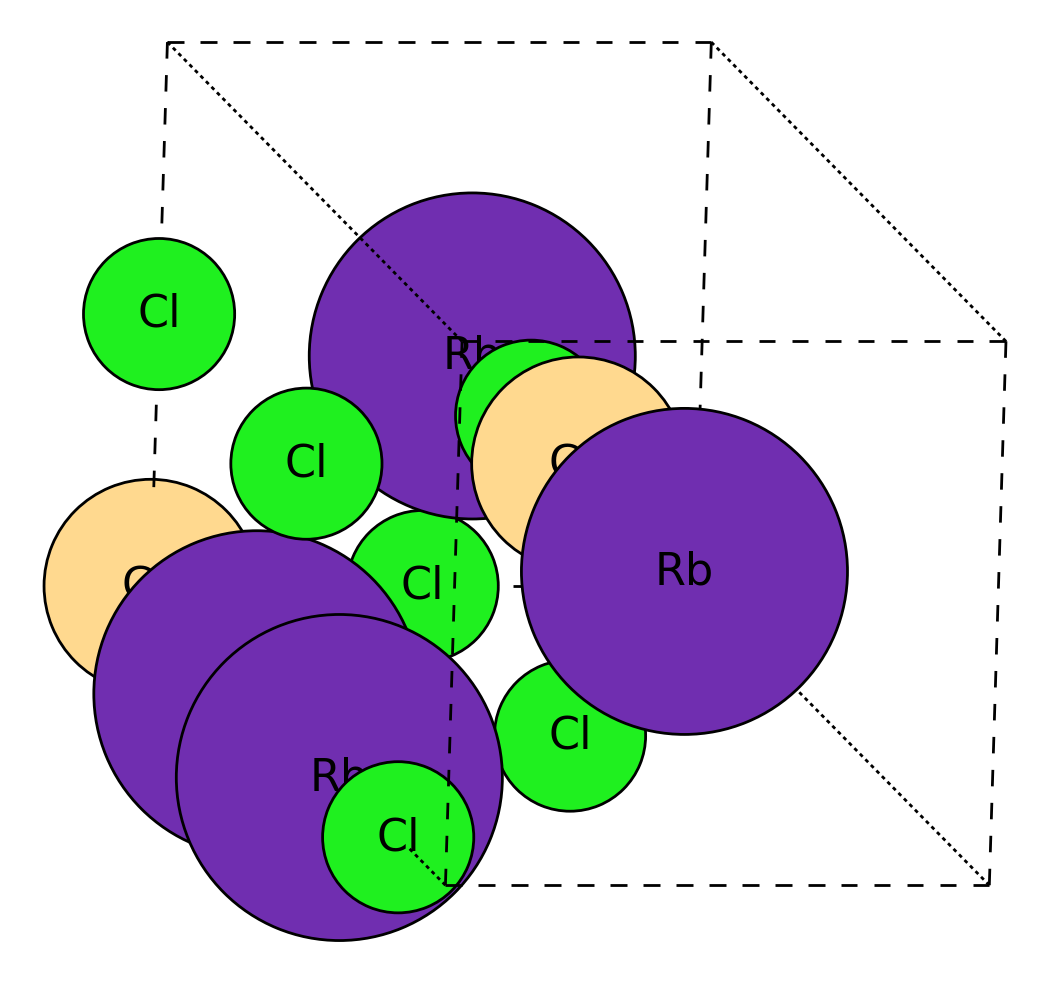}
	\end{minipage}
	\begin{minipage}{0.18\linewidth}
		\centering
            \text{mp-1235628}
            \text{\footnotesize{FE: -2.501 eV/atom}}
		\includegraphics[height=2.1cm]{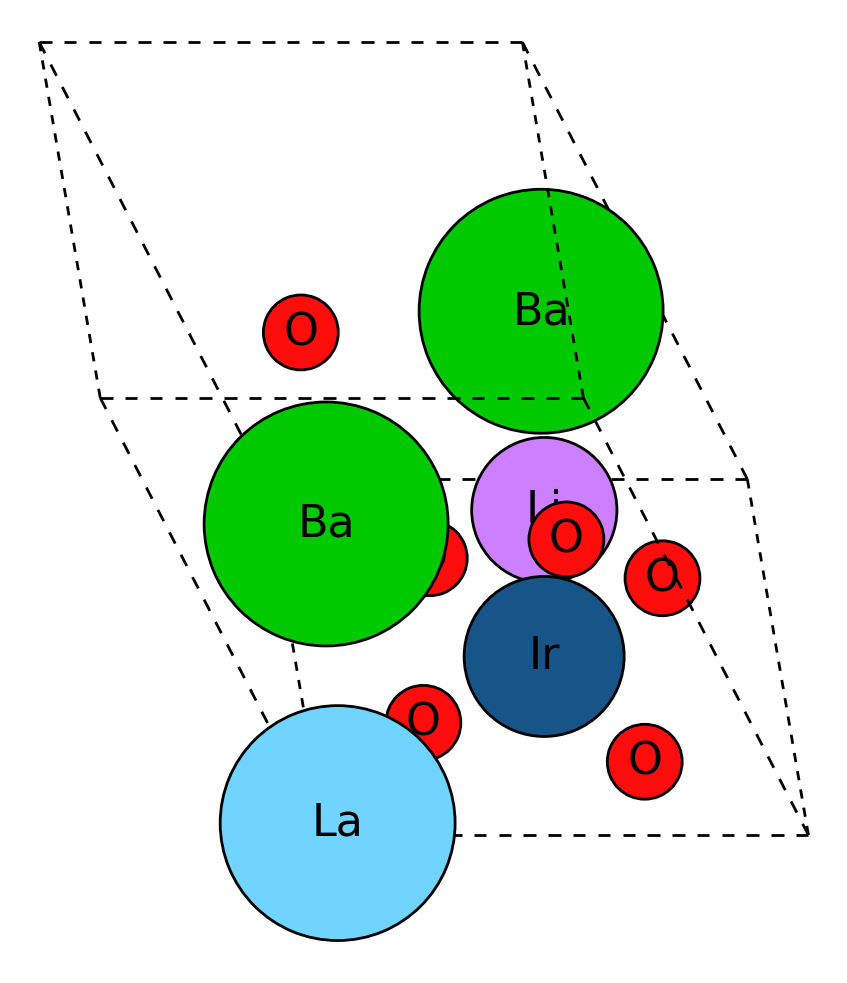}
	\end{minipage}
	\begin{minipage}{0.14\linewidth}
		\centering
            \text{mp-1187249}
            \text{\footnotesize{BG: 0.000 eV}}
		\includegraphics[height=2.1cm]{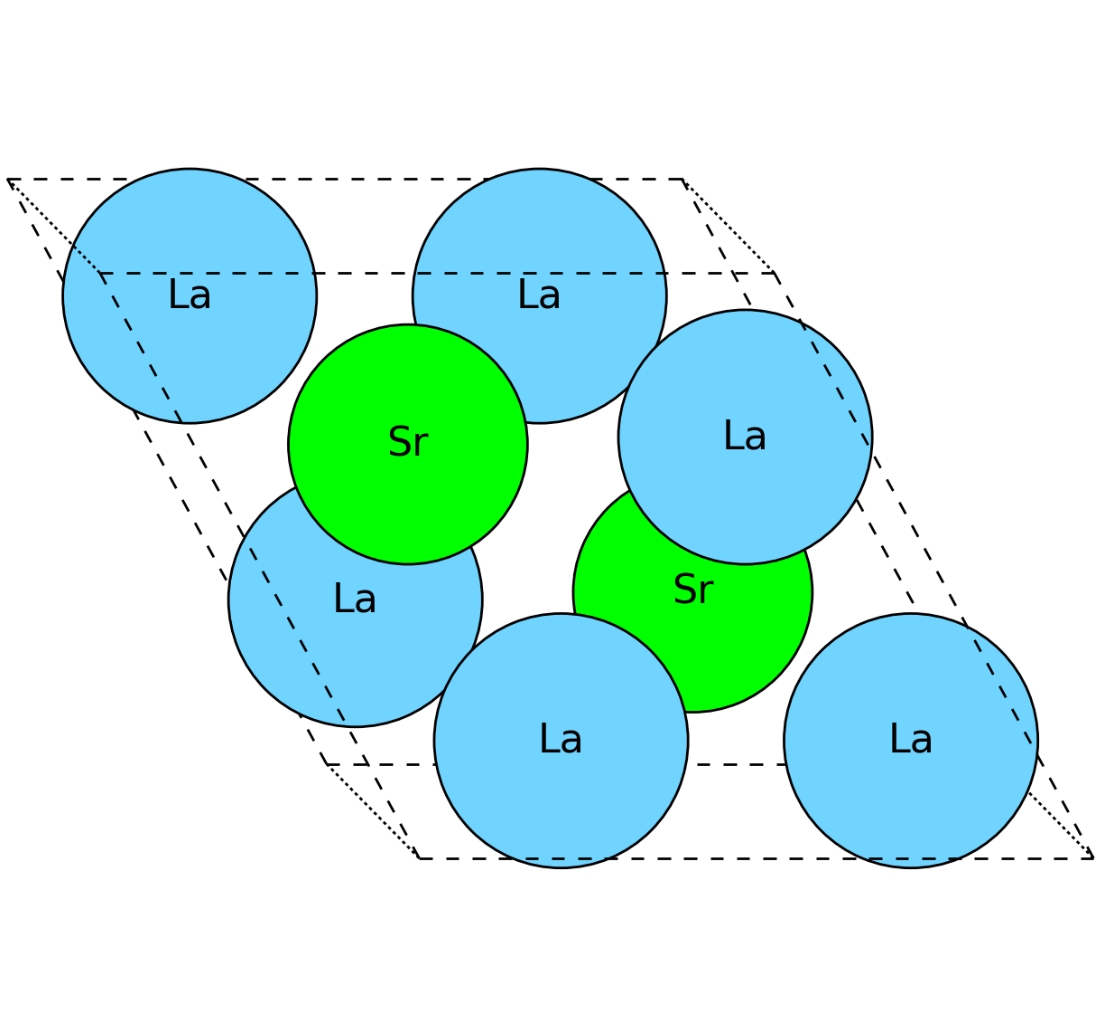}
	\end{minipage}
	
	\centering
	\begin{minipage}{0.18\linewidth}
		\centering
            \text{$\text{Dy}_{4}\text{In}_{2}\text{Ni}_{4}$}
            \text{\footnotesize{FE: -0.542 eV/atom}}
		\includegraphics[height=2.1cm]{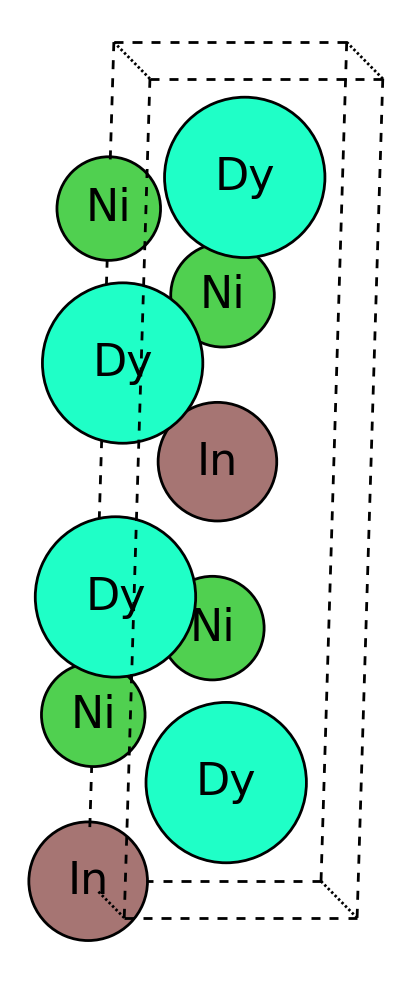}
	\end{minipage}
	\begin{minipage}{0.14\linewidth}
		\centering
            \text{$\text{Cs}_{4}\text{Na}_{2}\text{V}_{2}\text{O}_{8}$}
            \text{\footnotesize{BG: 3.576 eV}}
		\includegraphics[height=2.1cm]{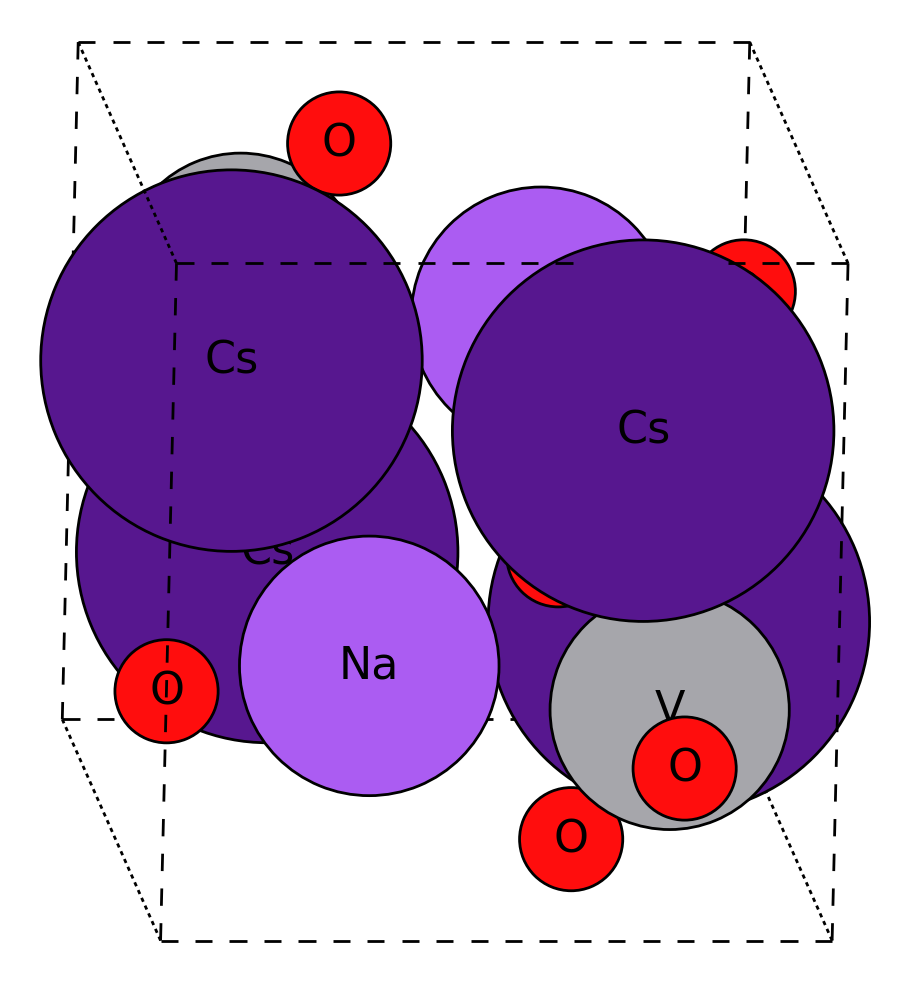}
	\end{minipage}
	\begin{minipage}{0.18\linewidth}
		\centering
            \text{$\text{K}_{2}\text{Gd}_{2}\text{Sn}_{2}\text{W}_{2}\text{O}_{12}$}
            \text{\footnotesize{FE: -2.209 eV/atom}}
		\includegraphics[height=2.1cm]{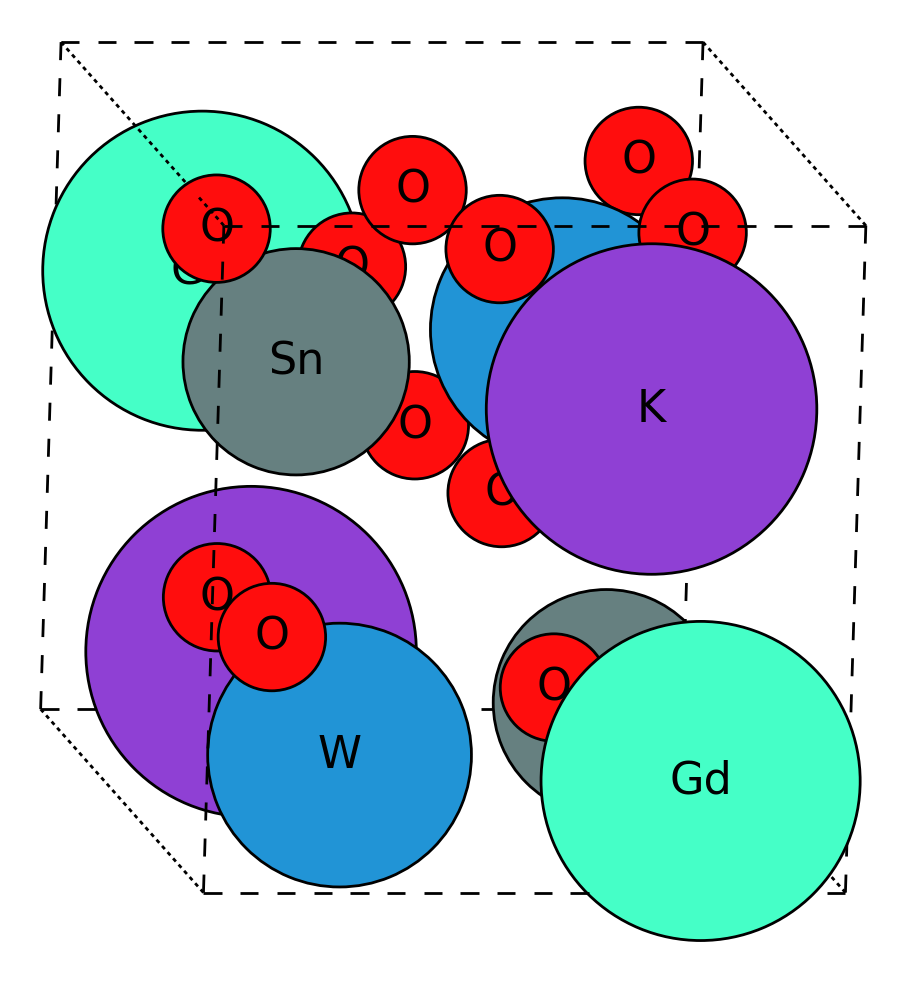}
	\end{minipage}
	\begin{minipage}{0.14\linewidth}
		\centering
            \text{$\text{Rb}_{4}\text{Cd}_{2}\text{Cl}_{8}$} 
            \text{\footnotesize{BG: 3.285 eV}}
		\includegraphics[height=2.1cm]{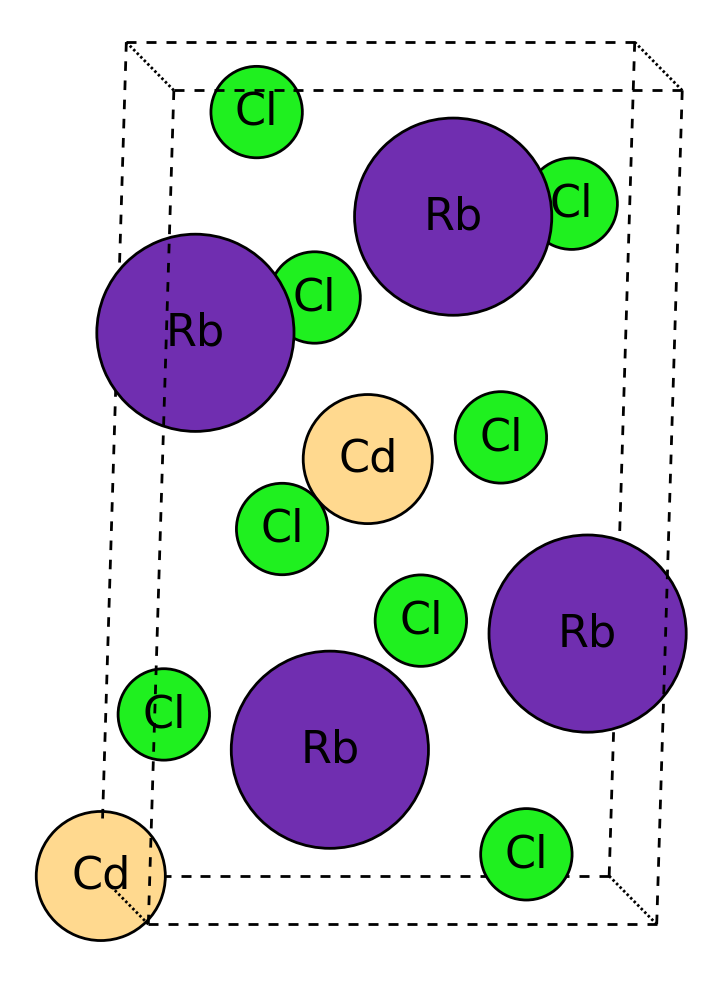}
	\end{minipage}
	\begin{minipage}{0.18\linewidth}
		\centering
            \text{$\text{Ba}_{2}\text{Li}_{1}\text{La}_{1}\text{Ir}_{1}\text{O}_{6}$}
            \text{\footnotesize{FE: -2.757 eV/atom}}
		\includegraphics[height=2.1cm]{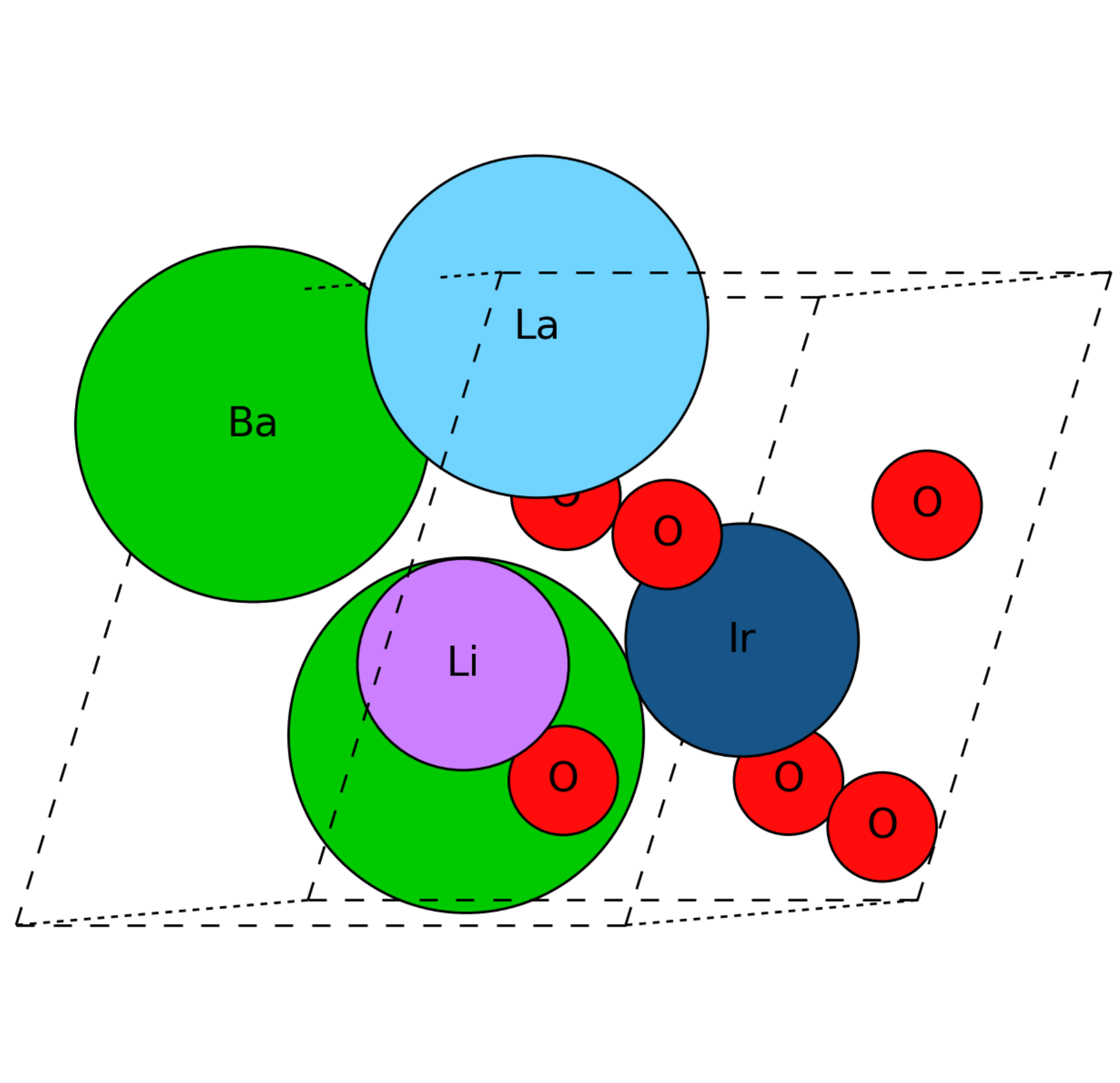}
	\end{minipage}
	\begin{minipage}{0.14\linewidth}
		\centering
            \text{$\text{Sr}_{2}\text{La}_{6}$}
            \text{\footnotesize{BG: 0.000 eV}}
		\includegraphics[height=2.1cm]{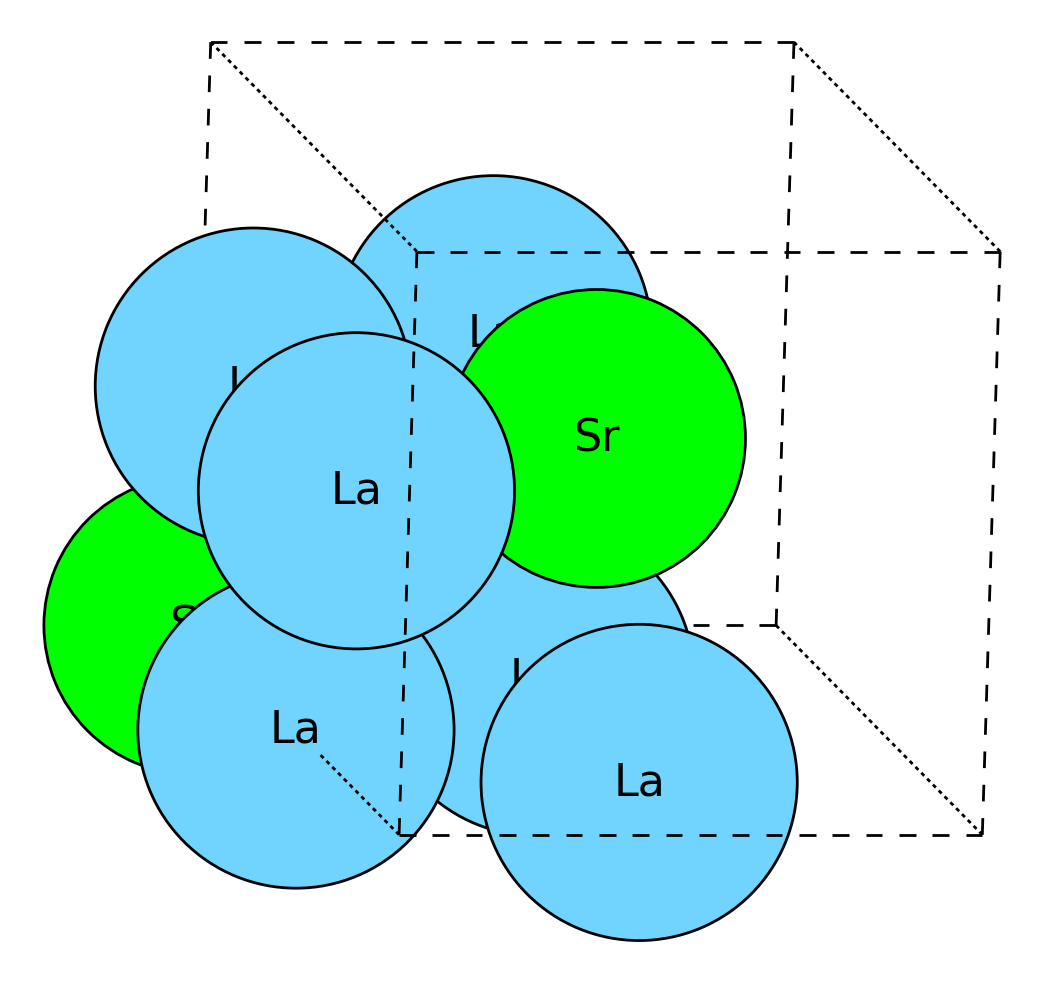}
	\end{minipage}
    \caption{Crystal structures generated by CrystalGF exhibit the different symmetry but similar material properties as those in the MP dataset.}
    \label{diff}
\end{figure*}

\section{Societal Impacts}\label{E}

Our proposed strictly constrained crystal structure generation framework represents a data-driven methodology that constitutes both a significant technological progress in materials science and a development with complex societal implications.

Traditional research and development of crystal materials are based on experimental trial and error, often over decades. Our structure generation framework addresses this challenge by analyzing extensive known structural data to generate target crystals within a broader symmetry space, thereby compressing material development cycles to weeks or even days while significantly reducing human resources, material costs, and time investment.

However, critical challenges remain. Sensitive data involved in advanced material development (e.g., military-grade material formulations) require the establishment of trusted data management platforms with robust data-sharing protocols and incentive mechanisms. Furthermore, data-driven crystal generation methods present emerging concerns including ambiguous copyright ownership and potential model hallucinations. These generated results should be cross-validated through rigorous experiments and expert review rather than being blindly relied upon.

\section{Limitations}\label{F}

We employ LLMs as the base models for our constraints generator, which significantly enhances the ability of our framework to learn complex relationships between material properties, chemical composition, and symmetry. However, this architecture increases computational costs for both fine-tuning and practical deployment. And through tens of thousands generation tests, we observe that in formula and property guided generation, approximately 0.069\% of cases exhibited failure in subsequent Wyckoff positions generation due to erroneous space group generation.

Furthermore, it can be found in Table \ref{CrystalGF-L-results} that while different generators show comparable overall performance within the same task, symmetry generation accuracy based on elemental composition showed significant inferiority compared to formula-based approaches. We attribute this discrepancy to the inherent structural information embedded in chemical formulas: Standard chemical formulas (e.g., Fe2O3 or TiO2) not only specify elemental composition but also implicitly encode potential bonding configurations and structural tendencies (e.g., spinel or rutile configurations), thereby providing more definitive classification clues. In contrast, mere elemental composition (e.g., Fe-O or Ti-O) lack constraints on atomic quantities and bonding patterns, forcing models to deduce complex symmetries from ambiguous inputs. Furthermore, the strong correlations between standard chemical formulas and space group classifications in existing material databases enable formula-based models to better capture structural patterns from training data. Conversely, models relying solely on elemental composition must contend with higher-dimensional uncertainties, as identical elemental composition may correspond to multiple compounds with distinct space groups, consequently diminishing generation accuracy.




\end{appendices}


\bibliography{sn-bibliography}

\end{document}